\documentclass{article}

     \PassOptionsToPackage{numbers, compress}{natbib}

     \usepackage[final]{neurips_2021}

\usepackage[utf8]{inputenc} 
\usepackage[T1]{fontenc}    
\usepackage{hyperref}       
\usepackage{url}            
\usepackage{booktabs}       
\usepackage{amsfonts, dsfont}       
\usepackage{nicefrac}       
\usepackage{microtype}      
\usepackage{xcolor}         
\usepackage{textcomp}

\usepackage{todonotes,nth}
\usepackage{graphicx}
\usepackage{cancel}
\usepackage{amssymb,amsmath,amsthm}
\usepackage{enumitem}
\usepackage{mathtools,algorithm,float}
\usepackage{dblfloatfix}
\usepackage{wrapfig}

\DeclareMathOperator*{\argmax}{arg\,max}
\DeclareMathOperator*{\argmin}{arg\,min}
\DeclareMathOperator{\KL}{KL}
\DeclareMathOperator{\kl}{kl}

\newtheorem{theorem}{Theorem}
\newtheorem*{theorem*}{Theorem}

\theoremstyle{definition}
\newtheorem{assumption}{Assumption}
\newtheorem{corollary}{Corollary}

\newtheorem{lemma}{Lemma}

\newtheorem*{Lemma*}{Lemma}

\newcommand\numberthis{\addtocounter{equation}{1}\tag{\theequation}}
\newcommand{\ie}{\textit{i.e.}}

\newcommand{\cf}{\textit{cf.}}

\title{Online Sign Identification:
	Minimization of the Number of Errors in Thresholding Bandits}

\author{%
  Reda Ouhamma\\
  Univ. Lille, Inria, CNRS,\\ Centrale Lille, UMR 9189 CRIStAL,\\ F-59000 Lille, France\\
  \texttt{reda.ouhamma@gmail.com} \\
  \And
  Rémy Degenne\\
  Univ. Lille, Inria, CNRS,\\ Centrale Lille, UMR 9189 CRIStAL,\\ F-59000 Lille, France\\
  \texttt{remy.degenne@inria.fr}  \\
   \AND
  Pierre Gaillard \\
  Univ. Grenoble Alpes, Inria, CNRS, \\
  Grenoble INP, LJK, 
  38000 Grenoble, France \\
  \texttt{pierre.gaillard@inria.fr} \\
  \And
  Vianney Perchet \\
  Crest, Ensae \& Criteo AI Lab \\
  \texttt{vianney.perchet@normalesup.org} 
}

\begin{document}

\maketitle

\begin{abstract}
In the fixed budget thresholding bandit problem, an algorithm sequentially allocates a budgeted number of samples to different distributions. It then  predicts  whether the mean of each distribution is larger or lower than a given threshold. We introduce a large family of algorithms (containing most existing relevant ones), inspired by the  Frank-Wolfe algorithm, and provide a thorough yet generic analysis of their performance. This allowed  us to construct new explicit algorithms, for a broad class of problems, whose losses are within a small constant factor of the   non-adaptive oracle ones.  Quite interestingly, we observed that  adaptive methods empirically greatly out-perform  non-adaptive oracles, an uncommon behavior  in standard online learning settings, such as regret minimization. We explain this surprising phenomenon on an insightful toy problem.

\end{abstract}

%
%
%
%
%
\section{Introduction and related work}

In a stochastic multi-armed bandit problem, a decision maker sequentially samples from different distributions in order to optimize a loss that depends on the unknown parameters of those distributions.
As a consequence, a tradeoff arises between gathering more samples from any possible distribution (to enhance the estimation of relevant parameters) and optimizing the allocation to minimize the final loss. We can distinguish two main categories of losses, focusing on ``exploitation'' vs ``exploration''.
The former directly depends on the whole allocation of samples and the typical example is regret minimization (we refer to the recent monographs \cite{lattimore2020bandit,bubeck2012regret,slivkins2019introduction} that cover this setting almost exhaustively).
The later is a bit different; after the budget of samples is exhausted, the algorithms must answer one or several ``questions'' (on the different distribution) and its loss is related to the number of mistakes made; the typical application being best-arm identification and variants \cite{audibert2010best,kaufmann2016complexity}.

We investigate a class of pure exploration problems, called ``thresholding bandit'' \cite{locatelli2016optimal,tao2019thresholding}.
The key property of this class is that a question is asked about each distribution, and the probability of making a mistake decreases with the total information gathered on that distribution solely. The typical question the algorithm must answer is ``is the mean of the distribution above or below some threshold?" (say, 0, for simplicity); giving the wrong answer can either incur a unit cost - independently from the distribution -, or a data-dependent cost (say, the distance to the threshold that represents the ``risk'' of that distribution). A typical application of  thresholding bandits is crowdsourcing \cite{chen2015statistical} where the objective is to distinguish workers with positive (vs.\ negative) efficiency; another one is bandit binary classification \cite{NEURIPS2019_e0b0f905}.

Some care must be taken when designing a performance criterion for a thresholding bandit problem, since any non-stupid algorithm will eventually answer all questions correctly (hence have a 0 loss) if it has enough samples. Furthermore, if distributions are sub-Gaussian (a rather mild assumption that we are going to make), the probability of making a single mistake decreases exponentially fast with the number of samples.
As a consequence, the focus must be on controlling the exponential decay constant.
We illustrate that issue on the unit cost problem described as follows. There are $K$ different $\sigma$-sub-Gaussian distributions; the mean of distribution $k$ is denoted by $\mu_k$ and the (variance-normalized) gap of distribution $k$ to the threshold 0 is denoted by $\Delta_k := \lvert \mu_k\rvert/\sqrt{2 \sigma^2}$. 
The algorithm has a budget of $T$ samples to (sequentially) allocate to those distributions and, based on the $N_{k,T}$ samples of distribution $k$, it must decide the sign of $\mu_k$; any mistake has a cost of one. We denote by $E_{k} \in \{0,1\}$ an indicator of a wrong sign prediction of $\mu_k$ after exhausting the budget of $T$ samples. The loss is then $L_T^{{1}} := \sum_k E_{k}$. It is not difficult to see that the expected number of mistakes could be of order $\sum_{k=1}^K \exp(- N_{k,T}\Delta_k^2)$~.

In particular, sampling evenly across distributions ($N_{k,T}=T/K$) gives $\mathbb{E}[L_T^1] \approx \sum_k\exp(-\frac{T}{K}\Delta_k^2)$, which has an exponential decay in $T$. However, this uniform allocation is far from being optimal in term of the exponential decay constant. Computing an (approximate) optimal fixed allocation in hindsight is not difficult: just optimize the upper-bound of $\mathbb{E}[L_T^1]$.
Since even the uniform allocation has a loss decaying exponentially, the performance of an algorithm should be measured not with respect to $\mathbb{E}[L_T^1]$ 
(see \cite{kaufmann2016complexity}) but rather in terms of $-\log(\mathbb{E}[L_T^1])/T$. The oracle that uses knowledge of the gaps $\Delta_k$ to optimize its fixed allocation verifies
\begin{align*}
\limsup_{T\to \infty}\frac{1}{T}\log(\mathbb{E}[L_T^1]) \le -\frac{1}{\sum_k 1/\Delta^2_k} \: .
\end{align*}

This unit cost framework has been investigated recently \cite{tao2019thresholding} with a simple yet effective algorithm called LSA (Logarithmic-Sample Algorithm) designed exclusively for this problem; it samples the distribution with the smallest current index defined as $\alpha N_{k,t} \hat{\Delta}_{k,t}^2 + \log N_{k,t}$, where $\hat{\Delta}_{k,t}$ is the empirical estimate of $\Delta_k$ and $\alpha$ is some parameter to be chosen. LSA is "optimal up to a constant", but the constant is unfortunately in the exponential decay, as it was proved that\footnote{See Remark 1 \cite{tao2019thresholding}. This bound implies that LSA - with the specified choice of $\alpha =0.1$ needs 16000 times more samples than the oracle to achieve the same performances.}
\begin{align*}
\limsup_{T\to \infty}\frac{1}{T}\log(\mathbb{E}[L_T^1]) \le -\frac{1}{16020}\frac{1}{\sum_k 1/\Delta^2_k}
\quad \text{ for LSA}.
\end{align*}
As we shall see, this result can be drastically improved with our more refined and general analysis (that implies choosing a totally different input parameter $\alpha = 1$ instead of $1/10$ as suggested originally).

\subsection{Contributions} 

We investigate the thresholding bandit problem with a weighted number of errors loss. Our contributions are twofold: 1) a generic method to design algorithms, with a generic proof, showing good performance on the weighted number of errors loss. 2) new lower-bounds and counter-intuitive results for the unit cost problem.

\textbf{A generic algorithm with performance guarantees~~} 
We propose a Frank-Wolfe inspired method to design bandit algorithms. We develop a proof technique to obtain loss bounds for the type of algorithms that our method produces, which we apply to the thresholding bandit with losses
\begin{equation}
	L_T = \sum_{k=1}^K a_k E_{k}
	\quad \text{or} \quad
	L_T^\Delta = \sum_{k=1}^K \Delta_k E_{k}
	\: ,
	\label{eq:weightedsumerrors}
\end{equation}
where $(a_k)_{k\in [K]}$ are known costs. 
The class of algorithms we analyze includes both LSA and APT (Anytime Parameter-free Thresholding) \cite{tao2019thresholding,locatelli2016optimal}. We obtain precise non-asymptotic loss bounds for $\mathbb{E}[L_T]$; for instance, we improve the original bound of LSA by a factor 4005  (and APT by a factor 8).
More importantly, we get a new algorithm whose expected error for the unit cost problem is within a factor 4 of the oracle.  We emphasize again than those ``constant'' factors are in the exponential (and are not mere multiplicative constants).

Interestingly, this class of algorithms are \emph{not} driven either by the ``optimism under uncertainty'' principle, a standard technique in multi-armed bandit \cite{auer2002finite} nor ``Explore-then-commit / Successive Elimination'' \cite{perchet2016batched,even2006action}. 

\textbf{New insights on the thresholding bandit problem~~}
First, the optimal allocation provided by the oracle of \cite{tao2019thresholding} in the unit cost problem has a M-shape (see Figure~\ref{fig:distribution_example}) because of two concurrent phenomena. On the one hand, the arms close to the threshold should not be pulled too much because their sign is difficult (if not impossible) to identify and it is a waste of budget.
On the other hand, the signs of the arms far from the threshold are quickly well estimated and therefore should not be chosen too often either. The middle arms are the ones that need to be pulled the most frequently. As $T$ gets larger, more and more budget is allocated to difficult arms.
In section~\ref{sub:number_of_pulls_lower_bound}, we provide a lower-bound that shows that this M shape is actually impossible to achieve for a sequential algorithm. Typically, the hollow inside of the M shape corresponds to arms whose sign cannot be well-estimated. In particular, it is not possible to distinguish arms that are very close to the threshold from the arms that are at the top of the M and should be pulled the most frequently according to the oracle. 

Our second insight is corroborated by numerical simulations in Section~\ref{sec:non_adaptive_algorithms}. We show empirically that our algorithms not only match but also surpass the optimal non-adaptive sampling of the oracle.
We conjecture that our algorithms take advantage of the chance due to noise that can move its estimate of the arm away from the threshold. In particular, when all the gaps $\Delta_k$ are equal, the non-adaptive optimal allocation should be uniform
, which is significantly outperformed by adaptive algorithms. This suggests that adaptivity is crucial for this problem and may inspire future research directions to the multi-armed bandit community in order to prove theoretical guarantees for such phenomena.

\subsection{Additional related work}


\textbf{Zero-one loss~~} Most of the literature on thresholding bandits \cite{locatelli2016optimal,mukherjee2017thresholding,cheshire2020influence} aims at minimizing the probability of making any sign error, i.e., minimizing the loss
\begin{equation}
L^*_T = \mathbb{I}\{\exists k \in [K], \ E_{k} = 1\} = \max_k E_{k} \label{eq:zero_one_error}.
\end{equation}
We already mentioned the algorithm APT \cite{locatelli2016optimal}, that gets an exponential decay of that loss (variants include variance estimation \cite{zhong2017asynchronous} and/or delayed feedbacks). Other algorithms exist, but based on the optimism principle \cite{katz2018feasible,mukherjee2017thresholding}. Unfortunately they suffer from a degraded exponential decay constant (by a factor bigger than 1000).

Another part of the literature focuses on the  fixed confidence framework, where the objective is to answer some questions with some fixed probability of mistake (and obviously with a minimal sample budget). For instance, an objective could be to return any arm above some threshold as soon as possible \cite{kano2019good,degenne2019pure}, or the one closest to the threshold \cite{garivier2017thresholding}, or just identifying that one arm is above that threshold \cite{kaufmann2018sequential}, or even to control false discovery rates and variants \cite{jamieson2018bandit,NEURIPS2019_e0b0f905}.

\textbf{Global loss, dynamic allocation and outliers detection~~}
The loss considered in thresholding bandits can be seen as a variant of a ``global loss'' (i.e., essentially non-linear) that has been extensively studied in the bandit literature \cite{agarwal2011,Agrawal:2014,Mannor}. However, the major difference is, again, that the optimal allocation is time dependent and that the loss converges exponentially fast to zero (no matter the algorithm). Similarly, Frank Wolfe algorithms have been introduced in this setting \cite{Berthet,Fontaine2}; even though our algorithms share some similarities, they are intrinsically different for the same reasons.

Similarly, the problem investigated could be seen as a special case of bandit resource allocations \cite{koopman,chen2015statistical,stochastic_ra,online_ra,fontaine2020adaptive} but where the loss is always decreasing with respect to the budget allocated per resource (hence again leading to a zero loss exponentially fast).

Finally the global objective of thresholding bandits is to obtain a synthetic view of how the means of distributions are spread on the real line (which ones are above/below some threshold). In that aspect, this problem sheds some similarities with outlier detection in multi-armed bandits \cite{NEURIPS2019_9b16759a,zhuang2017identifying,pmlr-v119-zhu20a}.

\section{Preliminaries}
\label{Preliminaries}
We describe here the weighted number of errors setting, in which an error on arm $k$ has a known cost $a_k>0$. The sum-of-gaps setting will be briefly investigated in section~\ref{ssub:the_sum_of_gaps_objective}.
The environment is composed of $K>1$ arms and an algorithm sequentially pulls them. After pulling  arm $k \in [K]$, it observes a sample from a distribution $\nu_k$ with mean $\mu_k$, and that sample is independent of past observations. The distribution $\nu_k$ is supposed $\sigma$-sub-Gaussian, that is
\begin{equation*}
\forall \lambda \in \mathbb{R}: \mathbb{E}_{X\sim \nu_k}\left[\exp(\lambda (X-\mu_k))\right]\leq \exp(\sigma^2 \lambda^2/2) \: .
\end{equation*}
The total number of rounds (and samples) $T$ is known in advance and called the horizon. After pulling $T$ arms, the task of the algorithm is to classify the arms depending on whether $\mu_k > \theta$ or not, where $\theta$ is a known threshold that we conveniently set to 0 (although it could be any other value, even different from arm to arm, without significant change to the analysis). Let $s_k \in \{-1, 1\}$ be the sign of $\mu_k - \theta$, equal to 1 iff $\mu_k - \theta>0$. The algorithm returns for all arms an estimated sign $\hat{s}_k \in \{-1, 1\}$. The objective is to minimize the expected weighted number of missclassified arms, where a mistake on arm $k$ has a known cost $a_k>0$,
\begin{equation}\label{def:loss_definition}
L_T = \sum_{k=1}^K a_k \mathbb{I}\{\hat{s}_k \ne s_k\} = \sum_{k=1}^K a_k E_{k} \: .
\end{equation} 
Note that the linear form of the loss is quite general: since $E_{k} \in \{0,1\}$, any separable loss $\sum_k f_k(E_{k})$ is the sum of a constant and $\sum_k a_k E_{k}$ for some costs $a_k$.

We conclude this description of the problem with notations used in the design of algorithms. Let $N_{k,t}$ and $\hat{\mu}_{k,t}=\frac{1}{N_{k,t}}\sum_{s=1}^t \mathbb{I}\{i_t=k\} X_t$ be the number of times the learner has pulled arm $k$ up to round $t$ (included) and the subsequent empirical mean of arm $k$ repectively. Define further $\hat{\Delta}_{k,t}=|\hat{\mu}_{k,t}-\theta|/\sqrt{2 \sigma^2}$ and $\Delta_{k}=|\mu_{k}-\theta|/\sqrt{2 \sigma^2}$, respectively the empirical and the true (variance-normalized) gap of arm $k$ to the threshold after $t$ rounds.


\subsection{The benchmarks: a lower bound and a non-adaptive oracle}
\label{subsec:Oracle}

Following the proof of \cite{tao2019thresholding} in a slightly more generic fashion (using exponential families with one parameter instead of Bernoulli distribution), we obtain a lower bound on the performance of any algorithm (see appendix~\ref{app:lower_bounds}) from which we get Theorem~\ref{thm:lower_bound_tao}.

\begin{theorem}{(Similar to Theorem 20 in \cite{tao2019thresholding})}\label{thm:lower_bound_tao}
Let $\left(\Delta_{1}, \ldots, \Delta_{K}\right)$ be a sequence of gaps. Then for any algorithm and time horizon $T \geq K$, there exists an instance in which all arms $k \in [K]$ have Gaussian distributions with variance $\sigma^2$ and mean in $\{\Delta_k, -\Delta_k\}$ such that
\begin{align*}
\mathbb{E}[L_T]
&\ge \frac{1}{4} \min_{\sum_k N_k = T} \sum_{k=1}^K a_k e^{- 4 N_k \Delta_k^2}
\: .
\end{align*}
\end{theorem}
We now deriving an optimal but unrealistic oracle, which requires prior knowledge of the gaps as input. 
Consider the algorithm that pulls each arm $N_{k,T}$ times, a number fixed in advance, then returns the sign of the empirical mean $\hat{\mu}_{k,T}$. Using Hoeffding's inequality, the expected loss verifies: 
\begin{equation}\label{ineq:Objective}
\mathbb{E}[L_T] = \sum_{k=1}^K a_k \mathbb{P} \left((\hat{\mu}_{k,T}- \theta)(\mu_k- \theta)<0\right) \leq \sum_{k=1}^K a_k e^{-N_{k,T} \Delta_k^2}
\end{equation}


We define the \emph{non-adaptive oracle} as the allocation $N_T$ which minimizes that upper bound. Its error probability has the same form as the lower bound of Theorem~\ref{thm:lower_bound_tao}, but has a different constant in the exponential (1 instead of 4). 
We can solve that minimization problem and make the error bound more explicit. To that end, suppose that the arms are ordered such that $a_1 \Delta_1^2 \le \ldots \le a_K \Delta_K^2$. 
There is a set $S = \{k_0, k_0+1,\ldots, K\}$ and a constant $C_S$ such that the oracle non-adaptive algorithm has $N_{k,T} = 0$ for $k \notin S$ and  $N_{k,T} = \left(C_S + \log(a_k \Delta_k^2)\right)/\Delta_k^2$ for $k \in S$ (see appendix~\ref{app:non_adaptive_oracle} for details). 
The expected loss of that non-adaptive oracle is
\begin{align}\label{eq:OptimalRegret}
\mathbb{E}[L_T]
\le \sum_{k \notin S} a_k
  + \sum_{k\in S} a_k \exp\left(- \frac{T + \sum_{j\in S} \frac{1}{\Delta_j^2}\log\left(\frac{a_k \Delta_k^2}{a_j \Delta_j^2}\right) }{\sum_{j\in S} \frac{1}{\Delta_j^2}} \right).
\end{align}

\begin{wrapfigure}{t}{0.37\textwidth}
\centering
\includegraphics[width=0.37\textwidth]{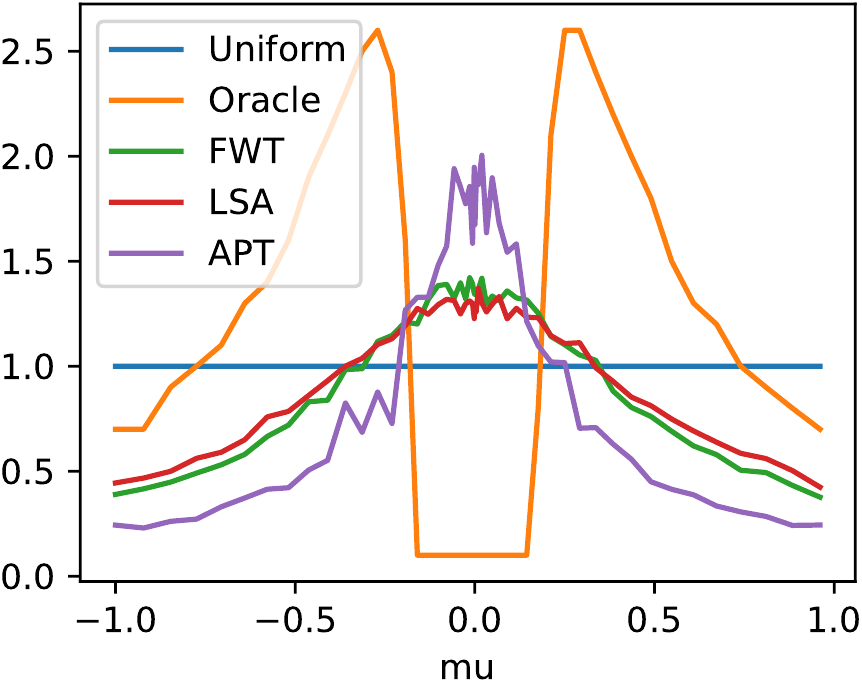} \vspace*{-15pt}
\caption{Optimal and empirical sampling distributions with respect to $\mu$.}\vspace*{-25pt}
\label{fig:distribution_example}
\end{wrapfigure}

The oracle is not pulling arms $1, \ldots, k_0-1$. These are the arms which are too close to the threshold (in a distance weighted by $a_k$) and thus too hard to classify to be worth trying. Giving up on those arms is not something that a non-oracle algorithm can do. 
Figure~\ref{fig:distribution_example} illustrates on an example ($\mu_k = (-1)^k (k/K)^2, k=1,\dots,50$, and $T=500$) the shape of the optimal allocation (arms near the threshold should not be drawn) as well as the empirical sampling distributions of several algorithms that pull all arms. In Appendix \ref{app:experiments}, we illustrate how this optimal allocation evolves with the horizon $T$.


\subsection{A good algorithm must pull all arms}
\label{sub:number_of_pulls_lower_bound}

We provide a new lower bound for the thresholding bandit with unit-cost problem,  to support the claim that it is not possible to avoid pulling the arms which are close to the threshold.
Consider the following 4 Gaussian bandit models (with variances 1) with means
\begin{align*}
\mu_{+\varepsilon} &= (\varepsilon, \ldots, \varepsilon, \mu_{K_0+1}, \ldots, \mu_K) \: , &
\mu_{+}' &= (\mu_{K_0+1}, \ldots, \mu_{K_0+1}, \mu_{K_0+1}, \ldots, \mu_K) \: ,\\
\mu_{-\varepsilon} &= (-\varepsilon, \ldots, -\varepsilon, \mu_{K_0+1}, \ldots, \mu_K) \: , &
\mu_{-}' &= (-\mu_{K_0+1}, \ldots, -\mu_{K_0+1}, \mu_{K_0+1}, \ldots, \mu_K) \: .
\end{align*}
%
where $0<\varepsilon < \mu_{K_0+1}\leq\ldots\leq \mu_K$, the value $\mu_{K_0+1}$ is large enough for the oracle to pull all arms on $\mu'_+$ and $\varepsilon \le \sqrt{\log(2)/(2 T)}$.

\begin{lemma}
If $\displaystyle\mathbb{E}_{\tilde{\mu}}[L_T] \le c_1 \min_{\sum_k N_k = T} \sum_k e^{-c_0 N_k \Delta_k^2}$ for constants $c_0, c_1$ on $\tilde{\mu} \in \{\mu'_+, \mu'_-\}$, then
\begin{align*}
\max_{\mu \in \{\mu_{+ \varepsilon}, \mu_{- \varepsilon}\}}\mathbb{E}_\mu\left[\sum_{k=1}^{K_0} N_{k,T} \right]
\ge \frac{1}{2(\mu_{K_0+1}-\varepsilon)^2} \left(c_0 \frac{T + H^{\log}}{H} + \log \frac{K_0}{32 c_1 H} \right)
\: .
\end{align*}
where $H = \frac{K_0}{\Delta_{K_0+1}^2} + \sum_{k=K_0+1}^K \frac{1}{\Delta_k^2}$ and $H^{\log} = \frac{K_0}{\Delta_{K_0+1}^2}\log\frac{1}{\Delta_{K_0+1}^2} + \sum_{k=K_0+1}^K \frac{1}{\Delta_k^2}\log\frac{1}{\Delta_k^2}$ .
\end{lemma}

The proof is postponed to Appendix~\ref{app:lower_bounds}. In a few words, if an algorithm has an expected loss close to the loss of the non-adaptive oracle, then it must pull linearly the arms which are close to the threshold.

\section{Algorithm and upper-bound}
\label{sec:OptimisticFranke-Wolfe}


We introduce and analyse a new class of algorithms for the thresholding bandit problem that we call \emph{index-based} algorithms.
That class unifies several existing algorithms, including APT~\citep{locatelli2016optimal} and LSA~\citep{tao2019thresholding}.
An index-based algorithm pulls the minimum of $K$ quantities, one for each arm, that each depends only on the rewards and pull counts of the respective arm (it does not change when pulling other arms). In particular, we consider algorithms for which the sampled arm is $i_{t+1} \in \argmin_{k \in [K]} F\big(N_{k,t}, N_{k,t} \hat \Delta_{k,t}^2; a_k\big)$ for a function $F:\mathbb{N} \times \mathbb{R}_+ \times \mathbb{R}_+^* \to \mathbb{R}$ that depends on the pull counts, the information about the sign and the weight of the arm.

\begin{algorithm}[!h]
{\bfseries Inputs}: an index function $F: \mathbb{N} \times \mathbb{R}_+ \times \mathbb{R}_+^* \to \mathbb{R}$; $a_1,\dots,a_K \in \mathbb{R}_+^*$; $\sigma > 0$; and $\theta \in \mathbb{R}$ \\[5pt]
For $t=1,\dots, T$ do

\begin{itemize}[parsep=0pt,itemsep=0pt, label={-},leftmargin=15pt]
	\item for all $k \in [K]$ define 
	\[
	N_{k,t-1} = \sum_{s=1}^{t-1} \mathbb{I}\{k = i_s\},\
	\hat \mu_{k,t-1} = \frac{1}{N_{k,t-1}} \sum_{s=1}^{t-1} \mathbb{I}\{k = i_s\} X_s, \
	\text{and} \
	\hat \Delta_{k,t-1}^2 = \frac{1}{2\sigma^2} \Big( \hat \mu_{k,t-1} - \theta \Big)^2
	\]
	\item pull $i_t \in \argmin_{k \in [K]} F\big(N_{k,t-1}, N_{k,t-1} \hat \Delta_{k,t-1}^2; a_k\big)$.
	\item observe $X_t \sim \nu_{i_t}$
\end{itemize}
Define $t_{\max} = \max_{t \in [T]} \min_{k \in [K]} F\big(N_{k,t}, N_{k,t} \hat \Delta_{k,t}^2; a_k\big)$ \\
Return for each $k \in [K]$ the sign $\hat s_k = \mathrm{sign}(\hat \mu_{k,t_{\max}} - \theta)$

\caption{Index-based algorithm for thresholding bandit}
\label{alg:index-based}
\end{algorithm}

After $T$ rounds, the algorithm recommends the sign of the arms at the round $t_{\max} \in [T]$ at which $\smash{\min_{k\in[K]}F\big(N_{k,t}, N_{k,t} \hat \Delta_{k,t}^2; a_k\big)}$ was maximal. This rule is used as opposed to returning the sign of all arms at time $T$ to facilitate the analysis, which is based on the observation that there is a small probability of error when all arms have high index. The time $t_{\max}$ should be close to $T$: in particular, only one arm is sampled (possibly several times) between $t_{\max}$ and $T$ (see Appendix~\ref{app:index_algorithms}).
In Sec.~\ref{sub:regret_upper_bound}, we provide a generic analysis for index-based algorithms satisfying  the assumption below. 

\begin{assumption}\label{ass:index_shape}
	The index function $F(n,x;a): \mathbb{N} \times \mathbb{R}_+ \times \mathbb{R}_+^* \to \mathbb{R}$ is non-decreasing in $n$ and $x$ and $\lim_{n \to +\infty} F(n, ny; a) = + \infty$ for all $y>0, a>0$.
\end{assumption}

Intuitively, algorithms that verify Assumption \ref{ass:index_shape} prefer pulling arms that were pulled the least (smallest $n$) and whose quantity of information about the sign ($n \smash{\hat{\Delta}_{k,n}^2}$) is small.
This class includes several algorithms from the thresholding bandits literature: APT \citep{locatelli2016optimal} for $F(n,x;a_k) = x$ and LSA \citep{tao2019thresholding} for $F(n,x;a_k) = x + \log(n)$ (these algorithms are only defined for $a_k = 1$). 
We now propose a generic method for designing an index-based algorithm.

\subsection{Frank-Wolfe for Thresholding bandits}
\label{sub:Frank-Wolfe}

Our strategy to minimize the expected loss is inspired by the Frank-Wolfe algorithm \citep{frank1956algorithm} and aims at controlling an upper-bound on the loss, such as the right hand side of Inequality \eqref{ineq:Objective}. Let's write that function as $\smash{B(N_T) = \sum_{k=1}^K a_k e^{- N_{k,T} \Delta_k^2}}$.
The high-level idea is to sequentially estimate its gradient and move to the minimizer of its linear approximation. 
If the gaps were known, we could compute at time $t+1$ the gradient of the bound with respect to $N_{t}$, $\nabla B(N_{t}) = (-a_k \Delta_k^2 e^{-N_{k,t} \Delta_k^2})_k$ and use the Frank-Wolfe algorithm. The algorithm would pull $i_{t+1} \in \argmin_u u^\top \nabla B(N_{t})$ for $u$ in the simplex, which is simply $\smash{\argmin_{k\in[K]}(-a_k \Delta_k^2 e^{-N_{k,t} \Delta_k^2})}$.
The gaps are however unknown. We therefore compute an estimate of the gaps $\hat{\Delta}_{k,t}$, with which we form the estimated gradient \
\[
\hat{\nabla} B(N_{t})_k = -a_k \hat{\Delta}_k^2 e^{-N_{k,t} \hat{\Delta}_{k,t}^2} = -\exp\left(-\left(N_{k,t} \hat{\Delta}_{k,t}^2  - \log (N_{k,t} \hat{\Delta}_{k,t}^2) + \log \Big(\frac{N_{k,t}}{a_k}\Big) \right) \right) \,.
\]
This gives a natural choice for the index function of our algorithm $F(n,x;a_k) = x - \log x +\log(n / a_k)$. However, the latter is decreasing in $x$ for $x \in (0,1)$, which in addition to violating Assumption~\ref{ass:index_shape},  may lead to instability in the initial phase when the gaps $\Delta_k$ are poorly estimated by $\smash{\hat \Delta_{k,n}}$.  We therefore propose a slight modification that preserves the asymptotic behavior of $F$ and we call the resulting algorithm FWT (Frank-Wolfe for Thresholding bandits):
\begin{equation}
	F(n,x;a_k) = \max\{x,1\} - \log(\max\{x,1\})+\log(n / a_k) \,. \tag{FWT} \label{def:FWT}
\end{equation}


\paragraph{Recovering APT}
\label{sub:relation_to_other_algorithms}

Using different upper-bounds $B$ on the expected loss may lead to different algorithms. In particular, we highlight a link between our Frank-Wolfe inspired method and the APT algorithm of \cite{locatelli2016optimal}, which was designed to minimize the loss $$L_T = \sum_{k=1}^K a_k \mathbb{I}\{\hat{s}_k \ne s_k\} = \sum_{k=1}^K a_k E_{k} \: .
$$ Following our method with the choice $\smash{B(N_t) = \max_{k\in[K]} e^{-N_{k,t} \Delta_k^2}}$ results in exactly the same sampling rule as the one of the APT algorithm (the recommendation rule differs slightly since we recommend the sign at $t_{\max}$ and not at $T$). Indeed, the derivative of $B$ with respect to $N_{k,t}$ is nonzero (and negative) if and only if $N_{k,t} \Delta_k^2 = \argmin_j  N_{j,t} \Delta_j^2$ (ignoring the case in which there are several argmins, for which the tie breaking can be arbitrary). 
This leads to the choice $F(n,x;a_k) = x$ in Algorithm~\ref{alg:index-based}, which then pulls 
$\smash{i_{t+1}= \argmin_{k\in[K]} N_{k,t} \hat{\Delta}_k^2}$. This is the sampling rule of APT. 

\subsection{Loss upper bound}
\label{sub:regret_upper_bound}

We provide a loss upper bound that is valid for all index-based algorithms that verify Assumption~\ref{ass:index_shape}. We then give a compact summary of the analysis outline and the resulting loss bounds.

\begin{theorem}\label{thm:expected_errors_bound}
	Let $K \geq 1$, $a_1,\dots,a_K >0$, $T\geq 1$, and $\sigma >0$. Let $F:\mathbb{N}\times \mathbb{R}\times \mathbb{R}_+^* \to \mathbb{R} $ that satisfies Assumption~\ref{ass:index_shape}. 
	Let $C_1,\ldots, C_K > \max_k F(0,0; a_k)$. For all $j,k \in [K]$, define
	\begin{itemize}[nosep]
		\item $t_j(C_k)$ a solution of the equation $F(t, t \Delta_j^2; a_j) = C_k$,
		\item $S_k \subseteq [K]$ and $t_{j,0}(C_k) \in \mathbb{R}_+$, a set and values such that for $i \notin S_k$, $\mathbb{P}\left(\exists n \le t_{i,0}(C_k), F(n,n\hat{\Delta}_{n,i}^2; a_i) \ge C\right)=1$.
	\end{itemize}
	Then the expected loss of Algorithm~\ref{alg:index-based} is upper-bounded as
	\begin{align*}
	\mathbb{E}[L_T^{\mathbb{A}}] \le &\sum_{k=1}^K a_k \left(e\cdot \exp\left(- 
	\frac{ \frac{1}{2}\left(T-\sum_{j \notin S_k} t_{j,0}(C_k)\right)- \sum_{j \in S_k} t_j(C_k)}{\sum_{j \in S_k} 1/\Delta_j^2}\right)+ T \cdot e^{- t_{k}(C_k) \Delta_k^2}\right)
	\: .
	\end{align*}
\end{theorem}
Refer to Appendix~\ref{app:Theorem1} for the proof. 
It is composed of two parts:
\begin{enumerate}
	\item First we establish that for any arm $j\in[K]$, with large probability, there is a time $\tau_j(C_k)$ such that $F(\tau_j(C_k),\tau_j(C_k)\hat{\Delta}_{\tau_j(C_k),j}; a_j)\ge C_k$. We prove that for all $j,k \in [K], \tau_j(C_k)$ has an exponential tail
	then use the fact that the algorithm pulls the minimal index to control the probability that the minimum never reaches $C_k$.
	\item We show that if an arm's index is large, then the probability of mistake on it is small.
\end{enumerate}
The times $t_j(C_k)$ of Theorem~\ref{thm:expected_errors_bound} are the smallest numbers of samples such that $t_j(C_k) \ge \tau_j(C_k)$ with high enough probability.
By determining those times, we derive explicit bounds for algorithms that verify Assumption~\ref{ass:index_shape}.
In particular we derive a bound for the variant of APT which returns the sign at the time $t_{\max}$ when the minimal index was maximal.
\begin{corollary} \label{cor:APT_upperbound}
	Suppose that for all $k \in [K]$, $a_k = 1$. For all $T\in \mathbb{N}^*$,
	\begin{align*}
	\mathbb{E}[L_T^{\text{APT}}]
	&\le 2 K\sqrt{e\cdot T} \cdot \exp \left( - \frac{1}{4}\frac{T}{\sum_{j = 1}^K 1/\Delta_j^2} \right)
	\: . 
	\end{align*}
\end{corollary}
Refer to Appendix~\ref{app:Examples} for the proof.
Since $\max_k E_k \le \sum_k E_k$, the bound of Corollary~\ref{cor:APT_upperbound} is also a bound on the zero-one loss, which we can compare to the result of \cite{locatelli2016optimal}. Our result shows a $1/4$ factor in the exponential instead of the worse $1/32$ constant of the original paper. 

\begin{figure}
\begin{center}
  \includegraphics[width=.35\textwidth]{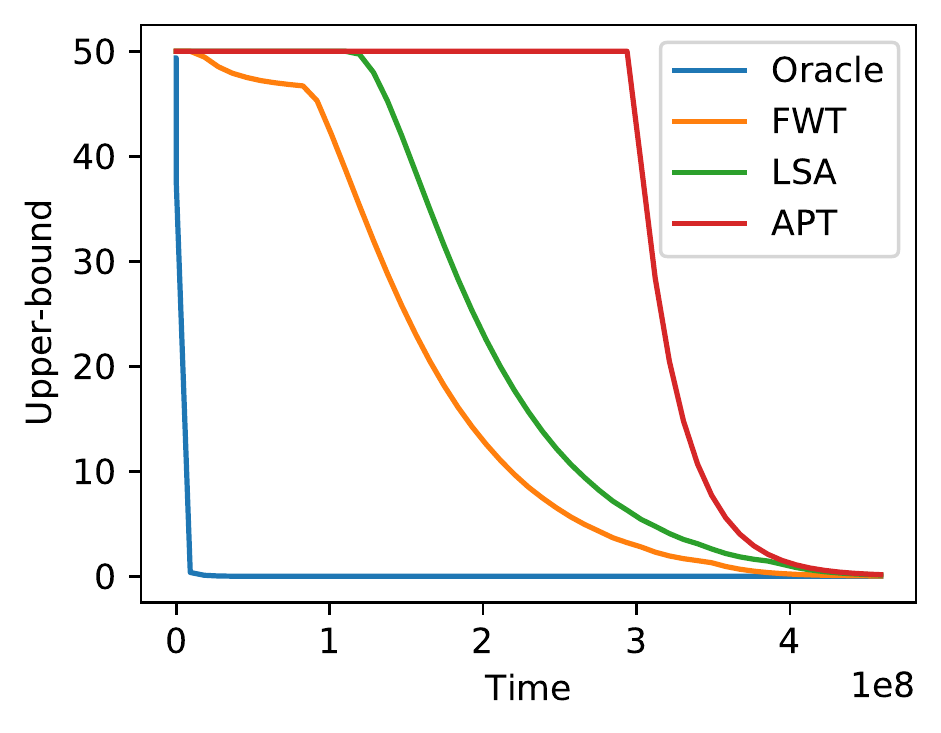}\qquad
  \includegraphics[width=.35\textwidth]{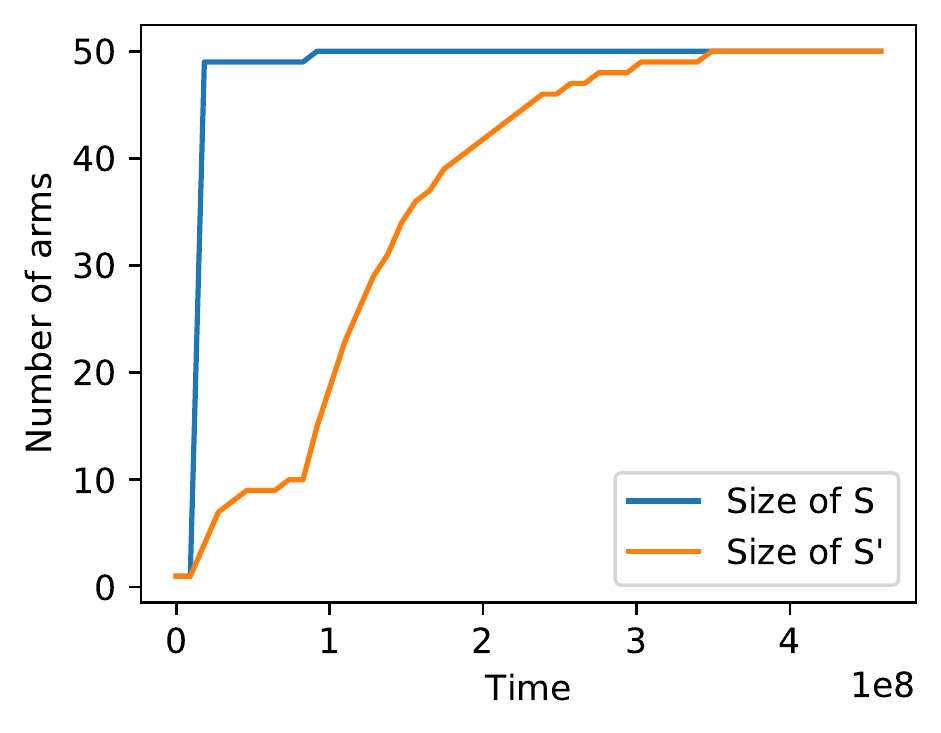} 
\end{center}
\caption{[left] Comparison of the upper-bounds of Corollaries~\ref{cor:APT_upperbound},~\ref{cor:FWTupperbound} and that of the optimal non-adaptive oracle of Eq.~\eqref{eq:OptimalRegret} (blue) when the gaps are of the form $\Delta_i = (i/K)^2$. [right] Evolution over time of the size of the optimal sets $S$ (blue) and $S'$ (orange) that minimize the bound of Corollary~\ref{cor:FWTupperbound}. }
\label{fig:upperbounds}
\end{figure}

\paragraph{LSA and FWT}
Theorem~\ref{thm:expected_errors_bound} applies to LSA and FWT with the following times:
\begin{itemize}[nosep]
	\item LSA:~~$t_j(C_k) = W(e^{C_k}\Delta_j^2)/\Delta_j^2$ and $t_{j,0}(C_k) = e^{C_k}$,
	\item FWT: $t_j(C_k) = \log(e^{C_k}a_j\Delta_j^2)/\Delta_j^2$ and $t_{j,0}(C_k) = a_j e^{C_k - 1}$ ,
\end{itemize}
where $W$ is the Lambert W function, which verifies $|W(x)-(\log x-\log\log x)| \le \log(1+1/e)$ for $x \ge e$. Therefore, for the two algorithms, the times $t_j(C_k)$ are close (equal up to the $\log\log$ terms in $W$), thus their bounds are close as well. Note that LSA is only defined for $a_j =1$ for all $j$.
In contrast to LSA, our bound for FWT has the notable property that, in the regime where $T \ge 2 \sum_{j=1}^K \frac{1}{\Delta_j^2} (2 + \log \frac{a_j\Delta_j^2 \max_i a_i \Delta_i^2}{(\min_k a_k \Delta_k^2)^2}- \log \frac{T}{e}) $, we recover the same exponent as in the non-adaptive oracle loss bound~\eqref{eq:OptimalRegret} (up to a factor $1/4$). Indeed we show that for such $T$
\begin{equation}
	\mathbb{E}[L_T^{\text{FWT}}] \le 2 \sqrt{e T} \sum_{k=1}^K a_k \exp \left( - \frac{1}{4} \frac{T + 2\sum_{j=1}^K \frac{1}{\Delta_j^2}\log \frac{a_k \Delta_k^2}{a_j\Delta_j^2}}{\sum_{j=1}^K 1/\Delta_j^2}\right)
	\: .\label{eq:large_T_bound_FWT}
\end{equation}
In the same regime of large $T$, the bound that we obtain for LSA is of the same order, but less explicit due to the function $W$. The latter is still impressive since the original theorem of \cite{tao2019thresholding} for LSA exhibits an exponent significantly looser, of order $\exp \left(-\frac{1}{16020} \frac{T}{\sum_{j=1}^K 1/\Delta_j^2}\right)$, \ie~ 4005-times worse than our bound.
We finally derive a bound for our newly introduced algorithm.
\begin{corollary}\label{cor:FWTupperbound}
	Let $S, S'$ be two sets with $S' \subseteq S \subseteq [K]$ and let $C \in \mathbb{R}$ be such that $C \ge 1 + \max\limits_{k \in S} \log \frac{1}{a_k \Delta_k^2}$. Then, for all $T\ge 1$
	\begin{align*}
	\mathbb{E}[L_T^{\text{FWT}}]
	&\le \sum_{k\notin S'} a_k
	+ e \sum_{k\in S'} a_k \exp\left(- \frac{\frac{1}{2}(T - \sum_{j \notin S} a_j e^{C-1}) + \sum_{j \in S} \frac{1}{\Delta_j^2}\log \frac{1}{a_j\Delta_j^2}}{\sum_{j \in S} 1/\Delta_j^2} + C \right)
	\\ &\qquad + T \sum_{k\in S'} a_k \exp \left( - C + \log (1/(a_k\Delta_k^2)) \right)
	\: .
	\end{align*}
\end{corollary}
Figure~\ref{fig:upperbounds} compares the upper-bounds of Corollary~\ref{cor:APT_upperbound} (APT), Theorem~\ref{thm:expected_errors_bound} (see also Equation~\eqref{eq:upper_bound_LSA} in the Appendix) (LSA), and Corollary~\ref{cor:FWTupperbound} (FWT) for the particular case $\Delta_i = (i/K)^2$ and $a_i = 1$, for $i = 1,\dots, K = 50$. See also Figure~\ref{fig:upperbounds_linear} in the supplementary material for $\Delta_i = i/K$. We can see that while the bounds of LSA, APT, and FWT are asymptotically similar, that of FWT starts to be significant for much smaller $T$. On the right, we can see the importance of the set $S'$ in Corollary~\ref{cor:FWTupperbound}: the bounds first ignores all the arms, and suffers a loss of 1 and then adds them one by one as soon as they can be classified. The bound derived in \cite{tao2019thresholding} for LSA is not represented on the figures, since it is still bigger than $K$ for the considered range of $T$.

\subsection{The sum-of-gaps objective}
\label{ssub:the_sum_of_gaps_objective}

We show that our method applies for the sum-of-gaps objective $\sum_{k = 1}^K \Delta_k E_k$. This is not a particular case of the setting discussed previously since $a_k$ was known to the algorithm, while $\Delta_k$ is unknown. It serves as a proof of concept for the extensibility of our method.
The index given by FWT in this setting is $F(n,x) = x' -\frac{3}{2}\log\left(x'\right)+\frac{3}{2}\log\left(n\right)$, where $x'=\max\left(x, \frac{3}{2}\right)$. We can then bound the sum-of-gaps loss using our generic analysis by proceeding similarly to Theorem~\eqref{thm:expected_errors_bound}.

\begin{corollary}\textbf{(FWT for the sum-of-gaps objective)}
	In the regime where $T\ge 2 \sum_{j=1}^k \frac{1}{\Delta_j^2}\left(3 + 3 \log\frac{\Delta_j \max_{i}\Delta_i}{(\min_{i}\Delta_i)^2}-\log\frac{T}{e}\right)$, we show that 
	\begin{equation*}
	\mathbb{E}[\sum_{k = 1}^K \Delta_k E_k] \le 2 \sqrt{e T} \sum_{k} \Delta_k \exp\left(- \frac{1}{2}\frac{\frac{T}{2} + \sum_{j} \frac{3}{2}\frac{1}{\Delta_j^2}\log \frac{\Delta_k^2}{\Delta_j^2}}{\sum_{j} 1/\Delta_j^2}\right).
	\end{equation*}
\end{corollary}
See Appendix~\ref{ap:sum_of_gaps} for the proof and for a different bound that is valid for all times $T$. This can be useful for applications in which errors are more tolerated for arms that are close to the threshold.

%
%
%

\section{Beating the oracle? The benefits of adaptivity.}
\label{sec:non_adaptive_algorithms}

We argue that in some situations adaptive algorithms can greatly outperform the non-adaptive oracle of Section~\ref{subsec:Oracle}, i.e., the cost of non-adaptivity can be much higher than the cost of learning. 
The algorithms in the family we considered are all adaptive in the sense that they adapt their drawing strategy as more information is observed, at the cost of learning the parameter $\mu_k$. We illustrate the benefits of adaptivity in the following toy example.
 
\paragraph{The ``optimal'' non-adaptive algorithm may be worse than adaptive algorithms.}
Consider the following parametric problem. An arm distribution is parametrized by $x \in \mathbb{R}$ and is supported on $\{0,x\}$; a sample of that distribution is equal to 0 or $x$, each with probability $1/2$. We assume that all arms have non-zero parameter and we will compute the optimal non-adaptive allocation. 
 
We make the convention that if an algorithm sees only zeros for one arm, it returns any sign with probability $1/2$. The error probability of a non-adaptive allocation $N_T^k$ for arm $k$ is half of the probability of seeing only zeros (since if anything else is observed, the arm can be classified with perfect accuracy). Hence the total error is
\begin{align*}
\mathbb{E}\big[L_T\big] = \frac{1}{2}\sum_{k=1}^K \frac{1}{2^{N_{k,T}}} \ge \frac{K}{2^{(T/K) + 1}} \: ,
\end{align*}
which is minimized with the uniform allocation: $N_{k,T} = \frac{T}{K}$ for all $k \in [K]$. 

Consider now an adaptive procedure that sample each arm in turn, but stops sampling an arm as soon at it sees a non-zero value. We crudely prove an upper bound for its number of errors, by remarking that it is zero if the algorithm classifies all arms correctly and smaller than $K$ otherwise.
The number of samples required to perfectly classify an arm follows a geometric distribution with parameter $1/2$. As a consequence, the number of required samples to classify all arms correctly follows a negative binomial NB($K, 1/2$). Let $Z$ be such a negative binomial random variable. The expected number of errors of the adaptive procedure is up to $K\mathbb{P}(Z > T)$. It then verifies
\begin{align*}
\mathbb{E}\big[L_T\big] \le 
K\mathbb{P}(Z>T)
\le K e^{-(\log(2)/2)T} \mathbb{E}e^{(\log(2)/2)Z}
=\frac{K}{2^{T/2}}\left(1 + \frac{1}{\sqrt{2}}\right)^K \: ,
\end{align*}
where the value $\log(2)/2$ is chosen for simplicity (in $[0, \log 2)$). In the regime where $T$ is large, this is of order $1/2^{T/2}$, which for $K>2$ is much smaller than $1/2^{T/K}$ for the uniform allocation.

\begin{figure}
  \begin{center}
  \includegraphics[width=.45\textwidth]{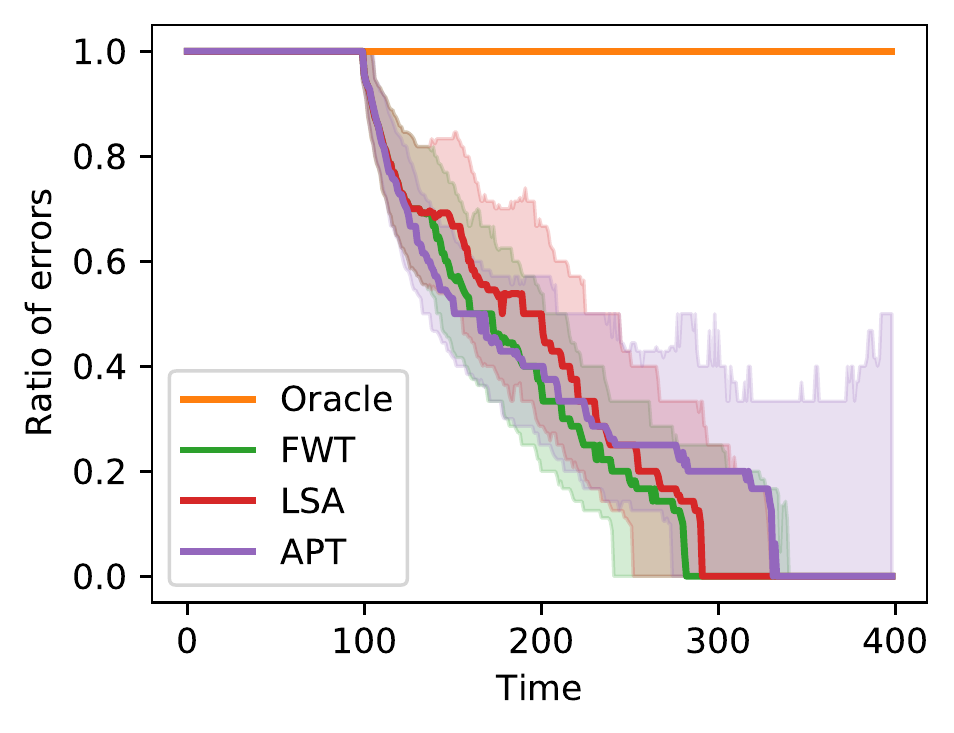} \quad 
  \includegraphics[width=.45\textwidth]{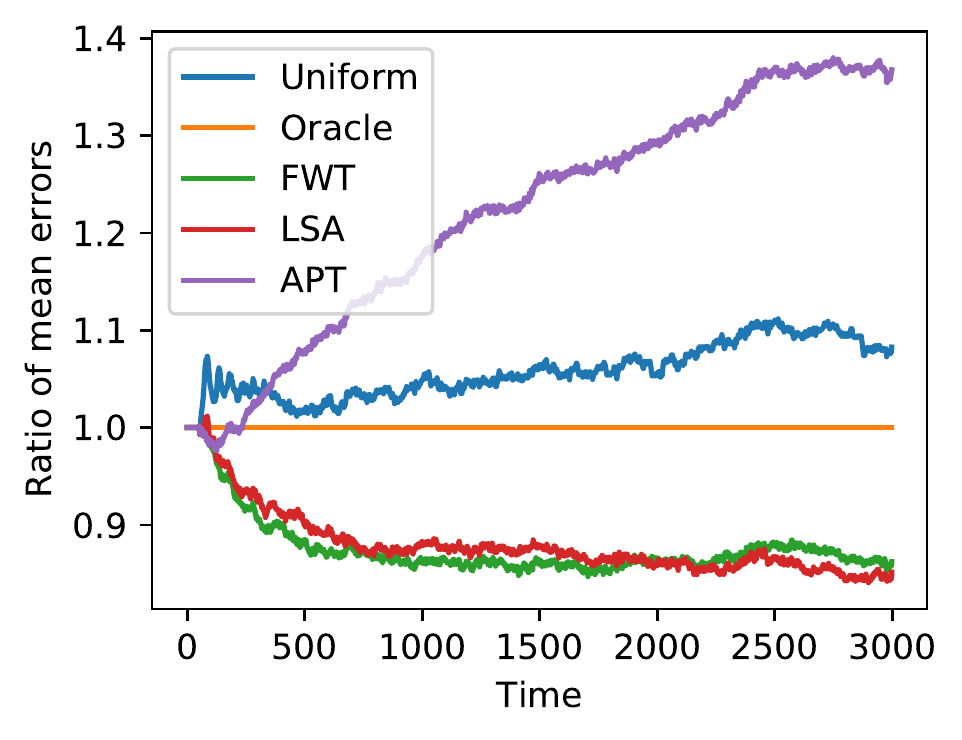} 
  \end{center}
  \caption{[left] Median (and $0.25$, $0.75$ empirical quantiles obtained on $500$ runs) of the ratio between the error suffered by each algorithm and that of the optimal non-adaptive oracle ($\mu_k = (-1)^k, k=1,\dots,100$).  [right] Ratio of the averaged errors (over $500$ runs) of each algorithm with that of the oracle ($\mu_k = (-1)^k(k/K)^2, k=1,\dots,50$).}
  \label{fig:adaptive-vs-nonadaptive}
\end{figure}

This toy example differs drastically from more realistic situations
, as one non-zero sample for an arm is sufficient to know the sign of the expectation perfectly. We therefore consider empirically more reasonable frameworks, closer to those analyzed in the paper: the distributions of $K$ arms are either $\mathcal{N}(1,1)$ or $\mathcal{N}(-1,1)$. Since all gaps $\Delta_i$ are equal, the optimal non-adaptive oracle is also the uniform sampling. 
The results are illustrated on the left part of Figure~\ref{fig:adaptive-vs-nonadaptive} and highlight the fact that all the adaptive algorithms considered (APT, LSA or FWT) drastically outperform the oracle. 
The right part of the figure shows the same phenomenon on another example in which the gaps are not constant. 
In particular, we can see that FWT and LSA have similar performance while APT (not designed for this purpose) generally suffers from a larger error. This result was corroborated by most of our experiments. We refer to Appendix~\ref{app:experiments} for more details.


\section*{Discussion}

An interesting research direction is to consider objective functions more general than~\eqref{eq:weightedsumerrors}. In particular, we believe that our approach can be generalized to losses of the form $\smash{L_T = \sum_{k=1}^K f(\Delta_k, E_k)}$ under certain regularity assumptions on $f$. Moreover, we focused on separable losses (hence linear wlog) and the index based algorithms we analyze reflect that separability. An obvious and intriguing direction for further work is to replace that assumption. One might for example want to design an algorithm that minimizes the probability of making more than a given number of mistakes. 

The fact that adaptive algorithms can beat non-adaptive oracles has already been observed empirically for fixed confidence identification \cite{simchowitz2017simulator,degenne2019non}, although only in cases where the non-adaptive oracle was worse only for small times and was still asymptotically optimal. The phenomenon we observe for fixed budget thresholding is much more significant and remains to be explained by theoretical arguments. Currently, the best theoretical bound for adaptive algorithms is still a factor $\nicefrac{1}{4}$ away in the exponent from the non-adaptive oracle bound.

\begin{ack}
V. Perchet acknowledges support from the French National Research Agency (ANR) under grant number \#ANR-19-CE23-0026 as well as the support grant, as well as from the grant  “Investissements d’Avenir” (LabEx Ecodec/ANR-11-LABX-0047)". R. Ouhamma also awknowledges support from Ecole polytechnique under the AMX funding. 
P. Gaillard and R. Degenne were supported by the French government under management of Agence Nationale
de la Recherche as part of the "Investissements d’avenir" program, reference ANR-19-P3IA-0001
(PRAIRIE 3IA Institute).

\end{ack}


\bibliography{references}

\begin{thebibliography}{38}
\providecommand{\natexlab}[1]{#1}
\providecommand{\url}[1]{\texttt{#1}}
\expandafter\ifx\csname urlstyle\endcsname\relax
  \providecommand{\doi}[1]{doi: #1}\else
  \providecommand{\doi}{doi: \begingroup \urlstyle{rm}\Url}\fi

\bibitem[Agarwal et~al.(2011)Agarwal, Foster, Hsu, Kakade, and
  Rakhlin]{agarwal2011}
Agarwal, A., Foster, D.~P., Hsu, D.~J., Kakade, S.~M., and Rakhlin, A.
\newblock Stochastic convex optimization with bandit feedback.
\newblock In Shawe-Taylor, J., Zemel, R.~S., Bartlett, P.~L., Pereira, F., and
  Weinberger, K.~Q. (eds.), \emph{Advances in Neural Information Processing
  Systems 24}, pp.\  1035--1043. Curran Associates, Inc., 2011.

\bibitem[Agrawal \& Devanur(2014)Agrawal and Devanur]{Agrawal:2014}
Agrawal, S. and Devanur, N.~R.
\newblock Bandits with concave rewards and convex knapsacks.
\newblock In \emph{Proceedings of the Fifteenth ACM Conference on Economics and
  Computation}, EC '14, pp.\  989--1006, New York, NY, USA, 2014. ACM.
\newblock ISBN 978-1-4503-2565-3.

\bibitem[Audibert et~al.(2010)Audibert, Bubeck, and Munos]{audibert2010best}
Audibert, J.-Y., Bubeck, S., and Munos, R.
\newblock Best arm identification in multi-armed bandits.
\newblock In \emph{COLT}, pp.\  41--53, 2010.

\bibitem[Auer et~al.(2002)Auer, Cesa-Bianchi, and Fischer]{auer2002finite}
Auer, P., Cesa-Bianchi, N., and Fischer, P.
\newblock Finite-time analysis of the multiarmed bandit problem.
\newblock \emph{Machine learning}, 47\penalty0 (2):\penalty0 235--256, 2002.

\bibitem[Berthet \& Perchet(2017)Berthet and Perchet]{Berthet}
Berthet, Q. and Perchet, V.
\newblock Fast rates for bandit optimization with upper-confidence frank-wolfe.
\newblock In Guyon, I., Luxburg, U.~V., Bengio, S., Wallach, H., Fergus, R.,
  Vishwanathan, S., and Garnett, R. (eds.), \emph{Advances in Neural
  Information Processing Systems 30}, pp.\  2225--2234. 2017.

\bibitem[Bubeck et~al.(2012)Bubeck, Cesa-Bianchi, et~al.]{bubeck2012regret}
Bubeck, S., Cesa-Bianchi, N., et~al.
\newblock Regret analysis of stochastic and nonstochastic multi-armed bandit
  problems.
\newblock \emph{Foundations and Trends{\textregistered} in Machine Learning},
  5\penalty0 (1):\penalty0 1--122, 2012.

\bibitem[Chen et~al.(2015)Chen, Lin, and Zhou]{chen2015statistical}
Chen, X., Lin, Q., and Zhou, D.
\newblock Statistical decision making for optimal budget allocation in crowd
  labeling.
\newblock \emph{The Journal of Machine Learning Research}, 16\penalty0
  (1):\penalty0 1--46, 2015.

\bibitem[Cheshire et~al.(2020)Cheshire, Menard, and
  Carpentier]{cheshire2020influence}
Cheshire, J., Menard, P., and Carpentier, A.
\newblock The influence of shape constraints on the thresholding bandit
  problem.
\newblock In \emph{Conference on Learning Theory}, pp.\  1228--1275. PMLR,
  2020.

\bibitem[Degenne \& Koolen(2019)Degenne and Koolen]{degenne2019pure}
Degenne, R. and Koolen, W.~M.
\newblock Pure exploration with multiple correct answers.
\newblock \emph{arXiv preprint arXiv:1902.03475}, 2019.

\bibitem[Degenne et~al.(2019)Degenne, Koolen, and M{\'e}nard]{degenne2019non}
Degenne, R., Koolen, W.~M., and M{\'e}nard, P.
\newblock Non-asymptotic pure exploration by solving games.
\newblock \emph{arXiv preprint arXiv:1906.10431}, 2019.

\bibitem[Devanur et~al.(2019)Devanur, Jain, Sivan, and Wilkens]{online_ra}
Devanur, N.~R., Jain, K., Sivan, B., and Wilkens, C.~A.
\newblock Near optimal online algorithms and fast approximation algorithms for
  resource allocation problems.
\newblock \emph{Journal of the ACM (JACM)}, 66\penalty0 (1):\penalty0 7, 2019.

\bibitem[Even-Dar et~al.(2006)Even-Dar, Mannor, Mansour, and
  Mahadevan]{even2006action}
Even-Dar, E., Mannor, S., Mansour, Y., and Mahadevan, S.
\newblock Action elimination and stopping conditions for the multi-armed bandit
  and reinforcement learning problems.
\newblock \emph{Journal of machine learning research}, 7\penalty0 (6), 2006.

\bibitem[Fontaine et~al.(2019)Fontaine, Berthet, and Perchet]{Fontaine2}
Fontaine, X., Berthet, Q., and Perchet, V.
\newblock Regularized contextual bandits.
\newblock In \emph{The 22nd International Conference on Artificial Intelligence
  and Statistics}, pp.\  2144--2153. PMLR, 2019.

\bibitem[Fontaine et~al.(2020)Fontaine, Mannor, and
  Perchet]{fontaine2020adaptive}
Fontaine, X., Mannor, S., and Perchet, V.
\newblock An adaptive stochastic optimization algorithm for resource
  allocation.
\newblock In \emph{Algorithmic Learning Theory}, pp.\  319--363. PMLR, 2020.

\bibitem[Frank \& Wolfe(1956)Frank and Wolfe]{frank1956algorithm}
Frank, M. and Wolfe, P.
\newblock An algorithm for quadratic programming.
\newblock \emph{Naval research logistics quarterly}, 3\penalty0 (1-2):\penalty0
  95--110, 1956.

\bibitem[Garivier et~al.(2017)Garivier, M{\'e}nard, Rossi, and
  Menard]{garivier2017thresholding}
Garivier, A., M{\'e}nard, P., Rossi, L., and Menard, P.
\newblock Thresholding bandit for dose-ranging: The impact of monotonicity.
\newblock \emph{arXiv preprint arXiv:1711.04454}, 2017.

\bibitem[Hoorfar \& Hassani(2008)Hoorfar and Hassani]{hoorfar2008inequalities}
Hoorfar, A. and Hassani, M.
\newblock Inequalities on the lambert w function and hyperpower function.
\newblock \emph{J. Inequal. Pure and Appl. Math}, 9\penalty0 (2):\penalty0
  5--9, 2008.

\bibitem[Jain \& Jamieson(2019)Jain and Jamieson]{NEURIPS2019_e0b0f905}
Jain, L. and Jamieson, K.~G.
\newblock A new perspective on pool-based active classification and
  false-discovery control.
\newblock In Wallach, H., Larochelle, H., Beygelzimer, A., d\textquotesingle
  Alch\'{e}-Buc, F., Fox, E., and Garnett, R. (eds.), \emph{Advances in Neural
  Information Processing Systems}, volume~32. Curran Associates, Inc., 2019.

\bibitem[Jamieson \& Jain(2018)Jamieson and Jain]{jamieson2018bandit}
Jamieson, K. and Jain, L.
\newblock A bandit approach to multiple testing with false discovery control.
\newblock \emph{arXiv preprint arXiv:1809.02235}, 2018.

\bibitem[Janson(2018)]{janson2018tail}
Janson, S.
\newblock Tail bounds for sums of geometric and exponential variables.
\newblock \emph{Statistics \& Probability Letters}, 135:\penalty0 1--6, 2018.

\bibitem[Kano et~al.(2019)Kano, Honda, Sakamaki, Matsuura, Nakamura, and
  Sugiyama]{kano2019good}
Kano, H., Honda, J., Sakamaki, K., Matsuura, K., Nakamura, A., and Sugiyama, M.
\newblock Good arm identification via bandit feedback.
\newblock \emph{Machine Learning}, 108\penalty0 (5):\penalty0 721--745, 2019.

\bibitem[Katariya et~al.(2019)Katariya, Tripathy, and
  Nowak]{NEURIPS2019_9b16759a}
Katariya, S., Tripathy, A., and Nowak, R.
\newblock Maxgap bandit: Adaptive algorithms for approximate ranking.
\newblock In Wallach, H., Larochelle, H., Beygelzimer, A., d\textquotesingle
  Alch\'{e}-Buc, F., Fox, E., and Garnett, R. (eds.), \emph{Advances in Neural
  Information Processing Systems}, volume~32. Curran Associates, Inc., 2019.

\bibitem[Katz-Samuels \& Scott(2018)Katz-Samuels and Scott]{katz2018feasible}
Katz-Samuels, J. and Scott, C.
\newblock Feasible arm identification.
\newblock In \emph{International Conference on Machine Learning}, pp.\
  2535--2543. PMLR, 2018.

\bibitem[Kaufmann et~al.(2016)Kaufmann, Capp{\'e}, and
  Garivier]{kaufmann2016complexity}
Kaufmann, E., Capp{\'e}, O., and Garivier, A.
\newblock On the complexity of best-arm identification in multi-armed bandit
  models.
\newblock \emph{The Journal of Machine Learning Research}, 17\penalty0
  (1):\penalty0 1--42, 2016.

\bibitem[Kaufmann et~al.(2018)Kaufmann, Koolen, and
  Garivier]{kaufmann2018sequential}
Kaufmann, E., Koolen, W., and Garivier, A.
\newblock Sequential test for the lowest mean: From thompson to murphy
  sampling.
\newblock \emph{arXiv preprint arXiv:1806.00973}, 2018.

\bibitem[Koopman(1953)]{koopman}
Koopman, B.~O.
\newblock The optimum distribution of effort.
\newblock \emph{Journal of the Operations Research Society of America},
  1\penalty0 (2):\penalty0 52--63, 1953.

\bibitem[Lattimore \& Szepesv{\'a}ri(2020)Lattimore and
  Szepesv{\'a}ri]{lattimore2020bandit}
Lattimore, T. and Szepesv{\'a}ri, C.
\newblock \emph{Bandit algorithms}.
\newblock Cambridge University Press, 2020.

\bibitem[Locatelli et~al.(2016)Locatelli, Gutzeit, and
  Carpentier]{locatelli2016optimal}
Locatelli, A., Gutzeit, M., and Carpentier, A.
\newblock An optimal algorithm for the thresholding bandit problem.
\newblock In \emph{International Conference on Machine Learning}, pp.\
  1690--1698. PMLR, 2016.

\bibitem[Mannor et~al.(2014)Mannor, Perchet, and Stoltz]{Mannor}
Mannor, S., Perchet, V., and Stoltz, G.
\newblock Approachability in unknown games: Online learning meets
  multi-objective optimization.
\newblock In \emph{Conference on Learning Theory}, pp.\  339--355. PMLR, 2014.

\bibitem[Mukherjee et~al.(2017)Mukherjee, Naveen, Sudarsanam, and
  Ravindran]{mukherjee2017thresholding}
Mukherjee, S., Naveen, K.~P., Sudarsanam, N., and Ravindran, B.
\newblock Thresholding bandits with augmented ucb.
\newblock \emph{arXiv preprint arXiv:1704.02281}, 2017.

\bibitem[Perchet et~al.(2016)Perchet, Rigollet, Chassang, Snowberg,
  et~al.]{perchet2016batched}
Perchet, V., Rigollet, P., Chassang, S., Snowberg, E., et~al.
\newblock Batched bandit problems.
\newblock \emph{Annals of Statistics}, 44\penalty0 (2):\penalty0 660--681,
  2016.

\bibitem[Salehi et~al.(2016)Salehi, Smith, Maciejewski, Siegel, Chong, Apodaca,
  Briceno, Renner, Shestak, Ladd, et~al.]{stochastic_ra}
Salehi, M.~A., Smith, J., Maciejewski, A.~A., Siegel, H.~J., Chong, E.~K.,
  Apodaca, J., Briceno, L.~D., Renner, T., Shestak, V., Ladd, J., et~al.
\newblock Stochastic-based robust dynamic resource allocation for independent
  tasks in a heterogeneous computing system.
\newblock \emph{Journal of Parallel and Distributed Computing}, 97:\penalty0
  96--111, 2016.

\bibitem[Simchowitz et~al.(2017)Simchowitz, Jamieson, and
  Recht]{simchowitz2017simulator}
Simchowitz, M., Jamieson, K., and Recht, B.
\newblock The simulator: Understanding adaptive sampling in the
  moderate-confidence regime.
\newblock In \emph{Conference on Learning Theory}, pp.\  1794--1834. PMLR,
  2017.

\bibitem[Slivkins et~al.(2019)]{slivkins2019introduction}
Slivkins, A. et~al.
\newblock Introduction to multi-armed bandits.
\newblock \emph{Foundations and Trends{\textregistered} in Machine Learning},
  12\penalty0 (1-2):\penalty0 1--286, 2019.

\bibitem[Tao et~al.(2019)Tao, Blanco, Peng, and Zhou]{tao2019thresholding}
Tao, C., Blanco, S., Peng, J., and Zhou, Y.
\newblock Thresholding bandit with optimal aggregate regret.
\newblock In \emph{Advances in Neural Information Processing Systems}, pp.\
  11664--11673, 2019.

\bibitem[Zhong et~al.(2017)Zhong, Huang, and Liu]{zhong2017asynchronous}
Zhong, J., Huang, Y., and Liu, J.
\newblock Asynchronous parallel empirical variance guided algorithms for the
  thresholding bandit problem.
\newblock \emph{arXiv preprint arXiv:1704.04567}, 2017.

\bibitem[Zhu et~al.(2020)Zhu, Katariya, and Nowak]{pmlr-v119-zhu20a}
Zhu, Y., Katariya, S., and Nowak, R.
\newblock Robust outlier arm identification.
\newblock In III, H.~D. and Singh, A. (eds.), \emph{Proceedings of the 37th
  International Conference on Machine Learning}, volume 119 of
  \emph{Proceedings of Machine Learning Research}, pp.\  11566--11575. PMLR,
  13--18 Jul 2020.

\bibitem[Zhuang et~al.(2017)Zhuang, Wang, and Wang]{zhuang2017identifying}
Zhuang, H., Wang, C., and Wang, Y.
\newblock Identifying outlier arms in multi-armed bandit.
\newblock In \emph{Proceedings of the 31st International Conference on Neural
  Information Processing Systems}, pp.\  5210--5219, 2017.

\end{thebibliography}
\bibliographystyle{neurips_2021}

%
%
%
%
%

\section*{Checklist}


\begin{enumerate}

\item For all authors...
\begin{enumerate}
  \item Do the main claims made in the abstract and introduction accurately reflect the paper's contributions and scope?
    \answerYes{}
  \item Did you describe the limitations of your work?
    \answerYes{}
  \item Did you discuss any potential negative societal impacts of your work?
    \answerNA{It is a purely theoretical paper}
  \item Have you read the ethics review guidelines and ensured that your paper conforms to them?
    \answerYes{}
\end{enumerate}

\item If you are including theoretical results...
\begin{enumerate}
  \item Did you state the full set of assumptions of all theoretical results?
    \answerYes{}
	\item Did you include complete proofs of all theoretical results?
    \answerYes{}
\end{enumerate}

\item If you ran experiments...
\begin{enumerate}
  \item Did you include the code, data, and instructions needed to reproduce the main experimental results (either in the supplemental material or as a URL)?
    \answerYes{See Appendix~\ref{app:experiments}.}
  \item Did you specify all the training details (e.g., data splits, hyperparameters, how they were chosen)?
    \answerYes{See the description of each figure.}
	\item Did you report error bars (e.g., with respect to the random seed after running experiments multiple times)?
    \answerYes{}
	\item Did you include the total amount of compute and the type of resources used (e.g., type of GPUs, internal cluster, or cloud provider)?
    \answerYes{See Appendix~\ref{app:experiments}.}
\end{enumerate}

\item If you are using existing assets (e.g., code, data, models) or curating/releasing new assets...
\begin{enumerate}
  \item If your work uses existing assets, did you cite the creators?
    \answerNA{}
  \item Did you mention the license of the assets?
    \answerNA{}
  \item Did you include any new assets either in the supplemental material or as a URL?
    \answerNA{}
  \item Did you discuss whether and how consent was obtained from people whose data you're using/curating?
    \answerNA{}
  \item Did you discuss whether the data you are using/curating contains personally identifiable information or offensive content?
    \answerNA{}
\end{enumerate}

\item If you used crowdsourcing or conducted research with human subjects...
\begin{enumerate}
  \item Did you include the full text of instructions given to participants and screenshots, if applicable?
    \answerNA{}
  \item Did you describe any potential participant risks, with links to Institutional Review Board (IRB) approvals, if applicable?
    \answerNA{}
  \item Did you include the estimated hourly wage paid to participants and the total amount spent on participant compensation?
    \answerNA{}
\end{enumerate}

\end{enumerate}


\appendix



\section{Lower Bounds}\label{app:lower_bounds}

We follow the method of \cite{tao2019thresholding}. Let $\hat{s}_k$ be the estimated sign of $\mu$. The expected loss on problem $\mu$ is
\begin{align*}
\mathbb{E}[L_T(\mu)]
= \sum_{k=1}^K a_k \mathbb{P}_\mu\{\hat{s}_k \ne s_k\}
\: .
\end{align*}

For each arm $k \in [K]$, we define two values $\mu_k, \tilde{\mu}_k \in \mathbb{R}$, with $\mu_k < \theta < \tilde{\mu}_k$. Let $\mu = (\mu_k)_{k\in [K]}$. For some fixed one-parameter exponential family, we denote by $\KL(a, b)$ the Kullback-Leibler divergence between distributions with mean $a$ and $b$. We recall the for Gaussians with variance $\sigma^2$, $\KL(a,b) = \frac{(a - b)^2}{2 \sigma^2}$.

\begin{theorem}\label{th:lb on flips}
For any algorithm,
\begin{align*}
\sup_{S \in \mathcal P([K])} \mathbb{E}[L_T(\mu_S)]
&\ge \frac{1}{4} \min_{N:\sum_k N_k = T} \sum_{k=1}^K a_k \exp \left( - N_k \max\{\KL(\mu_k, \tilde{\mu}_k),\KL(\tilde{\mu}_k, \mu_k)\} \right)
\end{align*}
In particular, for Gaussians with variance $\sigma^2$ and $\tilde{\mu}^k = \theta + (\theta - \mu^k)$,
\begin{align*}
\sup_{S \in \mathcal P([K])} \mathbb{E}[L_T(\mu_S)]
&\ge \frac{1}{4} \min_{N:\sum_k N_k = T} \sum_{k=1}^K a_k \exp \left( - 4 N_k \Delta_k^2 \right)
\: .
\end{align*}
\end{theorem}

\begin{proof}
Given a vector $\lambda \in \mathbb{R}^K$ with $\lambda_k \in \{\mu_k, \tilde{\mu}_k\}$ for all $k \in [K]$ and $S \subseteq [K]$, let $\lambda_S$ be such that $\lambda_{k,S} \in \{\mu_k, \tilde{\mu}_k\}$ and $\lambda_{k,S} \ne \lambda_k$ for $k\in S$ and $\lambda_{j,S} = \mu_j$ for $j\notin S$.

For $S \in \mathcal P([K])$, let $S \pm i$ be equal to $S\cup\{i\}$ if $i\notin S$ and to $S\setminus\{i\}$ otherwise. Also, we denote by $(s_k(\lambda))$ be the signs of $(\lambda_k)$. Then the following holds
\begin{align*}
\sup_{S \in \mathcal P([K])} \mathbb{E}[L_T(\mu_S)]
&\ge \frac{1}{2^K} \sum_{S \in \mathcal P([K])}\mathbb{E}[L_T(\mu_S)] 
\\
&= \frac{1}{2^K} \sum_{S \in \mathcal P([K])} \sum_{k=1}^K a_k\mathbb{P}_{\mu_S}\{\hat{s}_k \ne s_k(\mu_S)\} 
\\
&= \frac{1}{2^{K+1}} \sum_{S \in \mathcal P([K])} \sum_{k=1}^K a_k \mathbb{P}_{\mu_S}\{\hat{s}_k \ne s_k(\mu_S)\} + a_k\mathbb{P}_{\mu_{S \pm k}}\{\hat{s}_k \ne s_k(\mu_{S\pm k})\}
\\
&= \frac{1}{2^{K+1}} \sum_{S \in \mathcal P([K])} \sum_{k=1}^K a_k\mathbb{P}_{\mu_S}\{\hat{s}_k \ne s_k(\mu_S)\} + a_k\mathbb{P}_{\mu_{S \pm k}}\{\hat{s}_k = s_k(\mu_{S})\}.
\end{align*}
For each arm $k$, we can bound the sum of the two probabilities from below. Let $\mathcal E_{k,S} = \{\hat{s}_k \ne s_k(\mu_{S})\}$.
\begin{align*}
\mathbb{P}_{\mu_S}(\mathcal E_{k,S}) + \mathbb{P}_{\mu_{S \pm k}}(\overline{\mathcal E_{k,S}})
&\ge \frac{1}{2}\exp \left( - \mathbb{E}_{\mu_S}[N_{k,T}] \KL(\mu_{k,S}, \mu_{k,S\pm k}) \right) ,
\end{align*}
so that, when plugged back in the previous equation, we get
\begin{align*}
\sup_{S \in \mathcal P([K])} \mathbb{E}[L_T(\mu_S)]
&\ge \frac{1}{2^{K+1}} \sum_S \sum_{k=1}^K \frac{1}{2} a_k \exp \left( - \mathbb{E}_{\mu_S}[N_{k,T}] \KL(\mu_{k,S}, \mu_{k,S\pm k}) \right) 
\\
&\ge \frac{1}{4} \frac{1}{2^K}\sum_S \min_{N:\sum_k N_k = T} \sum_{k=1}^K a_k \exp \left( - N_k \KL(\mu_{k,S}, \mu_{k,S\pm k}) \right)
\\
&\ge \frac{1}{4} \min_{N:\sum_k N_k = T} \sum_{k=1}^K a_k \exp \left( - N_k \max\{\KL(\mu_k, \tilde{\mu}_k),\KL(\tilde{\mu}_k, \mu_k)\} \right)
\end{align*}
\end{proof}

\subsection{Lower bound on the number of pulls of arms close to zero}
\label{sub:lower_bound_on_the_number_of_pulls_of_arms_close_to_zero}

Consider a Gaussian bandit model (with variances $\sigma=1$) with vector of means $\mu_{+\varepsilon} = (\varepsilon, \ldots, \varepsilon, \mu_{K_0+1}, \ldots, \mu_K)$, in which arms $1,\ldots,K_0$ have mean $\varepsilon > 0$ and arms $K-K_0+1, \ldots, K$ have mean greater than $\varepsilon$. Let $\mu_{+}'$ be equal to $\mu_{+ \varepsilon}$ except that $\mu'_{+,j} = \mu_{K_0+1}$ for $j \in [K_0]$. We suppose that $\mu_{K_0+1}$ is large enough for the non-adaptive oracle to pull all arms on $\mu'_+$. We also define $\mu_{-\varepsilon} = (-\varepsilon, \ldots, -\varepsilon, \mu_{K_0+1}, \ldots, \mu_K)$ and $\mu'_- = (-\mu_{K_0+1}, \ldots, - \mu_{K_0+1}, \mu_{K_0+1}, \ldots, \mu_K)$.

\begin{lemma}
If an algorithm verifies $\mathbb{E}_{\tilde{\mu}}[L_T] \le c_1 \min_{N : \sum_k N_k = T} \sum_k e^{-c_0 N_k \tilde{\Delta}_k^2}$ for constants $c_0, c_1$ on all Gaussian problems with variance 1, for all mean vectors $\tilde{\mu}$ with gaps $\tilde{\Delta}$, then for $\varepsilon \le \sqrt{\log(2)/(2 T)}$ ,
\begin{align*}
\max_{\mu \in \{\mu_{+ \varepsilon}, \mu_{- \varepsilon}\}}\mathbb{E}_\mu\left[\sum_{k=1}^{K_0} N_{k,T} \right]
\ge \frac{1}{2(\mu_{K_0+1}-\varepsilon)^2} \left(c_0 \frac{T + H^{\log}}{H} + \log \frac{K_0}{32 c_1 H} \right)
\: .
\end{align*}
where $H = \frac{K_0}{\Delta_{K_0+1}^2} + \sum_{k=K_0+1}^K \frac{1}{\Delta_k^2}$ and $H^{\log} = \frac{K_0}{\Delta_{K_0+1}^2}\log\frac{1}{\Delta_{K_0+1}^2} + \sum_{k=K_0+1}^K \frac{1}{\Delta_k^2}\log\frac{1}{\Delta_k^2}$ .
\end{lemma}

\begin{proof}
We will prove that the number of pulls of arms $1,\ldots, K_0$ cannot be too small. Formally, let $n_\varepsilon = \sum_{k = 1}^{K_0} \mathbb{E}_{\mu_{+\varepsilon}} N_{k,T}$ be the expected number of pulls under $\mu_{+ \varepsilon}$ of the arms with mean $\varepsilon$. We aim at  showing that that number cannot be zero. We first prove that \begin{equation}\label{EQ:Lemma2}\mathbb{P}_{\mu_{+ \varepsilon}}(L_T > K_0/2) \ge \frac{1}{4}. \end{equation} This follows from the basic inequalities,
\begin{align*}
\mathbb{P}_{\mu_{+ \varepsilon}}(L_T({\mu_{+ \varepsilon}}) > K_0/2) + \mathbb{P}_{\mu_{-\varepsilon}}(L_T({\mu_{+ \varepsilon}}) \le K_0/2)
\ge \frac{1}{2}e^{-n_\varepsilon \KL(\varepsilon, -\varepsilon)}
\ge \frac{1}{2}e^{- 2 \varepsilon^2 T}
\end{align*}
In particular,  for $\varepsilon \le \sqrt{\frac{\log 2}{2 T}}$, $\max\{\mathbb{P}_{\mu_{+ \varepsilon}}(L_T({\mu_{+ \varepsilon}}) > K_0/2), \mathbb{P}_{\mu_{- \varepsilon}}(L_T({\mu_{-\varepsilon}}) > K_0/2)\} \ge \frac{1}{4}$, and either Inequality \ref{EQ:Lemma2} either holds for ${\mu_{+ \varepsilon}}$, or we just need to switch the role of $\varepsilon$ and $-\varepsilon$ in this proof. Suppose now that we  inequality \ref{EQ:Lemma2} holds for ${\mu_{+ \varepsilon}}$.

The Kullback-Leibler divergence between ${\mu_{+ \varepsilon}}$ and $\mu_+'$ is
\begin{align*}
\sum_k \mathbb{E}_{\mu}[N_{k,t}]\KL(\mu_{k,+ \varepsilon}, \mu'_{k,+})
&= \sum_{k=1}^{K_0} \mathbb{E}_{\mu}[N_{k,t}] \KL(\varepsilon, \mu_{k_0+1}) \ge \kl(\mathbb{P}_{\mu_{+ \varepsilon}}(L_T > K_0/2), \mathbb{P}_{\mu_+'}(L_T > K_0/2))
\\
&\ge \kl(\frac{1}{4}, \frac{2}{K_0}\mathbb{E}_{\mu_+'}[L_T]) \ge \frac{1}{4} \log \frac{K_0}{2\mathbb{E}_{\mu_+'}[L_T]} - \log 2
\: ,
\end{align*}
We have proved that $n_\varepsilon \ge \frac{1}{\KL(\varepsilon, \mu_{K_0+1})}(\frac{1}{4} \log \frac{K_0}{2\mathbb{E}_{\mu_+'}[L_T]} - \log 2)$ and the final result is obtained by using the explicit form for the bound on $\mathbb{E}_{\mu_+'}[L_T]$.
\end{proof}

\section{Non-adaptive oracle}
\label{app:non_adaptive_oracle}

The objective of this section is to explicit  the solution of
\begin{align*}
\min_{\sum_k N_k = T} \sum_{k} a_k e^{- N_k \Delta_k^2} \: .
\end{align*}

Introducing the Lagrange multiplier $\gamma \in \mathbb{R}$,  it is straightforward that the solution is such that all $N_k$ which are nonzero verify $\frac{\partial}{\partial N_k}(\sum_j a_j e^{- N_j \Delta_j^2}) = \gamma$. Then there exists a set $S$ and a constant $\gamma_S(T)>0$ for which $k \notin S \implies N_k = 0$ and for $k \in S$, $N_k \ne 0$ and
\begin{align*}
a_k \Delta_k^2 e^{- N_k \Delta_k^2} = \gamma_S(T) \: .
\end{align*}
That is, $N_k = \frac{1}{\Delta_k^2}(\gamma_S(T) + \log(a_k \Delta_k^2))$ .

We remark that $k \in S$ iff $\frac{1}{\Delta_k^2}(\gamma_S(T) + \log(a_k \Delta_k^2))>0$, which then implies that if $a_1 \Delta_1^2 \le \ldots \le a_K \Delta_K^2$, then $S = \{k_0, k_0+1, \ldots, K\}$ for some $k_0 \in [K]$.

Using the condition $\sum_k N_k = T$ to determine $\gamma_S(T)$, we get
\begin{align*}
\sum_{k = k_0}^K \frac{1}{\Delta_k^2}(\gamma_S(T) + \log(a_k \Delta_k^2)) = T
\quad \implies \quad
\gamma_S(T) = \frac{T + \sum_{k = k_0}^K \frac{1}{\Delta_k^2}\log\frac{1}{a_k \Delta_k^2}}{\sum_{k = k_0}^K \frac{1}{\Delta_k^2}}
\: .
\end{align*}
Finally, we can characterize $k_0$. Notice that $k \in S$ iff $\frac{1}{\Delta_k^2}(\gamma_S(T) + \log(a_k \Delta_k^2))>0$, i.e. iff
\begin{align*}
\frac{1}{\Delta_k^2}\left(\frac{T + \sum_{j = k_0}^K \frac{1}{\Delta_j^2}\log\frac{1}{a_j \Delta_j^2}}{\sum_{j = k_0}^K \frac{1}{\Delta_j^2}} + \log(a_k \Delta_k^2) \right)
> 0 
\quad \Leftrightarrow \quad
T > \sum_{j = k_0}^K \frac{1}{\Delta_j^2}\log\frac{a_j \Delta_j^2}{a_k \Delta_k^2}
\: .
\end{align*}
Finally, let $H_k = \sum_{j = k+1}^K \frac{1}{\Delta_j^2}\log\frac{a_j \Delta_j^2}{a_k \Delta_k^2}$, with $H_0 = +\infty$ and $H_K = 0$. Then $k_0$ is the unique element of $[K]$ such that $H_{k_0} < T \le H_{k_0 - 1}$.

\section{Properties of index-based algorithms}\label{app:index_algorithms}

An algorithm is index-based if, at any round $t$, it pulls $k_t = \argmin_k I_{N_{k,t-1}}^k$ where the index $I_{N_{k,t}}^k$ depends only on the number of pulls and on rewards of arm $k$. That index does not change when other arms are pulled.

For $C \ge 0$, let $\mathcal F_C \triangleq \{ \exists T' \le T, \forall k \in [K], I_{N_{k,T'}}^k \ge C \}$ be the event that at some time before $T$, all arm indices are above a value $C$. And let $\tau_k(C) = \min \{n | I_{n}^k \ge C\}$ be the minimal number of pulls of arm $k$ such that its index becomes greater than $C$.

We start with two immediate remarks about index-based algorithms.
\begin{lemma}\label{lem:min_I_ge_C_of_pulled_again}
	If $I_{N_{j,t}}^j \ge C$, then at the next time $t'$ when an index-based algorithm pulls arm $j$, it necessarily holds that $\min_k I_{N_{t'-1}^k}^k \ge C$.
\end{lemma}
\begin{lemma}\label{lem:all_N_ge_tau_of_min_I_ge_C}
	If $\min_k I_{N_{k,t}}^k \ge C$ then for all $k$, by definition of $\tau_k(C)$, $N_{k,t} \ge \tau_k(C)$.
\end{lemma}

This next lemma explicits $\mathcal F_C$ using $(\tau_k(C))_{k \in [K]}$.

\begin{lemma}\label{lem:F_eq_sum_tau_le_T}
	An index-based algorithm verifies $\mathcal F_C = \{\sum_k \tau_k(C) \le T\}$.
\end{lemma}
\begin{proof}
	We first prove  the inclusion $\mathcal F_C \subseteq \{\sum_k \tau_k(C) \le T\}$. At the time $T'$ defined in $\mathcal F_C$, it holds $\min_k I_{N_{T'}}^k \ge C$. The results then follows from lemma~\ref{lem:all_N_ge_tau_of_min_I_ge_C}: $\sum_k \tau_k(C) \le \sum_k N_{k,T'} = T' \le T$.\\
	
	We now prove $\{\sum_k \tau_k(C) \le T\} \subseteq \mathcal F_C$.
	If there is no $j$ with $N_{j,T} > \tau_j(C)$, we have $T = \sum_k N_{k,T} \le \sum_k \tau_k(C) \le T$. Hence there is equality and we have $N_{k,T} = \tau_k(C)$ for all $k$ and $\mathcal F_C$ is true for $T'=T$.

	If there is some $j$ such that $N_{j,T} > \tau_j(C)$, then after the time at which arm $j$ was pulled $\tau_j(C)$ times it verified $I_{N_{j,t}}^j \ge C$. Arm $j$ is again pulled at least once at some time $t'$, and at that time we have by lemmas~\ref{lem:min_I_ge_C_of_pulled_again} that for all $k$, $I_{N_{k,t'-1}}^k \ge I_{N_{j,t'-1}}^j = I_{N_{j,t}}^j \ge C$. Stated otherwise, the event $\mathcal F_C$ happens.
\end{proof}

\begin{lemma}
Let $t_{\max} = \argmax_{t \in [T]} \min_{k\in [K]} I_{N_{k,t}}^k$. Then for all arms except at most one, $N_{k,t_{\max}}= N_{k,T}$.
\end{lemma}
\begin{proof}
The algorithm switches arm only if the index of the pulled arm becomes strictly greater than the minimal index of the others. As a consequence, the value of the minimal index at times of arm changes is increasing. If two or more arms are pulled since $t_{\max}$, there is an arm change later than $t_{\max}$ and the minimal index value at that time is higher than at $t_{\max}$. This is a contradiction.
\end{proof}


%


\section{Loss upper bound}\label{app:Theorem1}

Notation: we analyze index functions slightly more general than $F(n, x; a_k)$. Each arm has a potentially different index function $F_k(n, x)$.

Consider then algorithms which obey -a slightly more generic- Assumption~\ref{ass:index_shape}, \ie~whose index can be written as $I_n^k = F_k(n,n \hat{\Delta}_{n,k}^2)$, where each $F_k$ is non-decreasing in both variables, and $\lim_{n \to +\infty} F_k(n, ny) = + \infty$ for $y>0$. And recall the theorem we aim at  proving.
\begin{theorem*}
	Let $K \geq 1$, $a_1,\dots,a_K >0$, $T\geq 1$. For $k \in [K]$, let $F_k:\mathbb{N}\times \mathbb{R} \to \mathbb{R}$  satisfying Assumption~\ref{ass:index_shape}. 
	Let $C_1,\ldots, C_K > \max_k F_k(0,0)$. For all $j,k \in [K]$, define
	\begin{itemize}[nosep]
		\item $t_j(C_k)$ a solution of the equation $F_j(t, t \Delta_j^2) = C_k$,
		\item $S_k \subseteq [K]$ and $t_{j,0}(C_k) \in \mathbb{R}_+$, a set and values such that for $i \notin S_k$, $\mathbb{P}\left(\exists n \le t_{i,0}(C_k), F_i(n,n\hat{\Delta}_{n,i}^2) \ge C\right)=1$.
	\end{itemize}
	Then the expected loss of Algorithm~\ref{alg:index-based} is upper-bounded as
	\begin{align*}
	\mathbb{E}[L_T^{\mathbb{A}}] \le &\sum_{k=1}^K a_k \left(e\cdot \exp\left(- 
	\frac{ \frac{1}{2}\left(T-\sum_{j \notin S_k} t_{j,0}(C_k)\right)- \sum_{j \in S_k} t_j(C_k)}{\sum_{j \in S_k} 1/\Delta_j^2}\right)+ T \cdot e^{- t_{k}(C_k) \Delta_k^2}\right)
	\: .
	\end{align*}
\end{theorem*}

\begin{proof}
Set $(C_k)_{1\le k\le K} \in \mathbb{R}^+$, so that it immediately follows
\begin{align*}
\mathbb{E} [L_T]
&= \sum_k a_k \mathbb{P}(\hat{s}_k \ne s_k)
\\
&\le \sum_{k=1}^K a_k \mathbb{P}(\hat{s}_k \ne s_k \land \exists T' \in [T], I_{N_{T'}}^k \ge C_k) + a_k \mathbb{P}(\forall T' \in [T], I_{N_{T'}}^k < C_k)
\\
&\le  \sum_{k=1}^K a_k \mathbb{P}(\hat{s}_k \ne s_k \land \exists T' \in [T], I_{N_{T'}}^k \ge C_k) + a_k \mathbb{P}(\overline{\mathcal F_{C_k}})
\: ,
\end{align*}
where $\overline{\mathcal{E}}$ stands for the complement of an event $\mathcal{E}$. 
The proof then proceeds in two steps, which are proved in subsections \ref{sub:with_large_probability_all_arms_are_well_explored} and \ref{sub:when_arm_well_explored}, in order to control both probabilities introduced above:
\begin{enumerate}[nosep]
	\item Lemma~\ref{lem:sum_tau_tail}: with large probability, there is some time $t$ for which the index $I_{N_{k,t}}^k$ is large for all $k \in [K]$ (and all arms are well explored)
	
	\item Theorem~\ref{lem:proba_error_tail}: if $I_{N_{k,t}}^k$ is large, then there is a small probability of mistake for arm $k$.
\end{enumerate}
\end{proof}

\subsection{With large probability, all arms are well explored}
\label{sub:with_large_probability_all_arms_are_well_explored}

\paragraph{Technical tools used in this section}
First, we state a simple but useful lemma.
\begin{lemma}\label{def:Delta_lower_confidence_bound}
	Let $\Delta_{n,k}^{\delta} = \Delta_k - \sqrt{\frac{\log(1/\delta)}{n}}$. For all $\delta \in (0,1)$ such that $\Delta_{n,k}^{\delta} \ge 0$, Hoeffding's inequality implies that $\mathbb{P}(\hat{\Delta}_{n,k} < \Delta_{n,k}^{\delta}) \le \delta$.
\end{lemma}

The next lemma will be used to bound the sum of exponentially tailed distributions.
\begin{lemma}[\cite{janson2018tail}]\label{lem:sum_exponentials}
	Let $Z_1, \ldots, Z_K$ be independent random variables and $a_1, \ldots, a_K \in \mathbb{R}^+$ be such that for all $k \in [K]$ and $x \in \mathbb{R}^+$, $\mathbb{P}(Z_k \ge x) \le e^{- a_k x}$. Then for all $\lambda \ge 0$,
	\begin{align*}
	\mathbb{P}(\sum_k Z_k \ge \lambda \sum_k \frac{1}{a_k}) \le e^{1 - \lambda} \: .
	\end{align*}
\end{lemma}

\begin{corollary}\label{cor:sum_exponentials}
	Let $Y_1, \ldots, Y_K$ be independent random variables and $y_1, \ldots, y_K \in \mathbb{R}, \  a_1, \ldots, a_K \in \mathbb{R}^+$ be such that for all $k \in [K]$ and $x \in \mathbb{R}^+$, $\mathbb{P}(Y_k \ge y_k + x) \le e^{- a_k x}$. Then for all $x \ge \sum_k y_k$,
	\begin{align*}
	\mathbb{P}(\sum_k Y_k \ge x) \le e \times \exp\left( -\frac{x - \sum_k y_k}{ \sum_k 1/a_k} \right) \: .
	\end{align*}
\end{corollary}
The corollary is a direct application of Lemma~\ref{lem:sum_exponentials} to $Z_k = Y_k - y_k$.

\paragraph{Main proof}

We know from Lemma~\ref{lem:F_eq_sum_tau_le_T} that any index based algorithm verifies $\mathcal F_C = \{\sum_k \tau_k(C) \le T\}$. Hence, to prove that $\mathcal{F}_C$ happens with great probability, it suffices to show that $\sum_k \tau_k(C)$ has an exponential tail.

We derive a bound on $\mathbb{P}(\sum_k \tau_k(C) > T)$. To that end, we bound individually for each arm $\mathbb{P}(\tau_k(C) \ge t_k + x)$ for some $t_k$ to be defined and $x\ge 0$, and conclude by Corollary~\ref{cor:sum_exponentials}.

\begin{theorem}
Under Assumption~\ref{ass:index_shape}, the algorithm verifies, for all $x \ge 0$,
\begin{align*}
\mathbb{P}(\sqrt{\tau_k(C)} > \sqrt{t_k(C)} + x) \le \exp(- \Delta_k^2 x^2) \: ,
\end{align*}
with $t_k(C)$ solution to $F_k(t, t \Delta_k^2) = C$, if such a solution exists. Otherwise, if $C < F_k(0,0)$ and no solution exists, $\tau_k(C) = 0$ with probability 1.
\end{theorem}
\begin{proof}
We first bound $\mathbb{P}(\tau_k(C) > t_k + x)$, where $t_k$ is chosen later, and $x \ge 0$.
\begin{align*}
\mathbb{P}(\tau_k(C) > t_k + x)
= \mathbb{P}(\forall n \le t_k + x,\  I_n^k < C)
&\le \mathbb{P}(I_{t_k + x}^k < C)
\\
&= \mathbb{P}(F_k(t_k+x, (t_k+x)\hat{\Delta}_k^2) < C)
\end{align*}

First, by monotonicity of $F_k$, this probability equals zero if $C \le F_k(t_k+x, 0)$.
If $C > F_k(t_k+x, 0)$, we define $\delta_{t_k+x,k,C}$ such that $\Delta_{t_k+x,k}^{\delta_{t_k+x, k, C}} = \inf\{ \Delta\ge 0 \ | \ C < F_k(t_k+x, (t_k+x)\Delta^2) \}$. 
In the following, we write $F_k(n, \cdot)^{-1}(C) := \inf\{ x \ | \ C \le F_k(n,x) \}$. If $F_k(n, \cdot)$ is increasing, this is its inverse, but we only suppose that $F_k(n, \cdot)$ is non-decreasing. Note that $x < F_k(n, \cdot)^{-1}(C)$ implies that  $F_k(n, x) < C$. With that definition, $(t_k+x)(\Delta_{t_k+x,k}^{\delta_{t_k+x, k, C}})^2 = F_k(t_k+x, \cdot)^{-1}(C)$. As a consequence, we get
\begin{align*}
\mathbb{P}(\tau_k(C) > t_k + x)
\le \mathbb{P}(I_{t_k + x}^k < C)
&= \mathbb{P}(F_k(t_k + x, (t_k+x)\hat{\Delta}_{t_k+x,k}^2) < C)
\\
&\le \mathbb{P}((t_k+x)\hat{\Delta}_{t_k+x,k}^2 < F_k(t_k+x, \cdot)^{-1}(C))
\\
&= \mathbb{P}(\hat{\Delta}_{t_k+x,k}^2 < (\Delta_{t_k+x,k}^{\delta_{t_k+x, k, C}})^2)
\\
&\le \delta_{t_k+x,k,C} \: ,
\end{align*}
where by definition,
\begin{align*}
\delta_{t_k+x,k,C}
&= \exp\left( - (t_k+x)(\Delta_k - \Delta_{t_k+x,k}^{\delta_{t_k+x, k, C}}) \right)
\\
&= \exp\left( - \left( \sqrt{\Delta_k^2(t_k+x)} - \sqrt{F_k(t_k+x, \cdot)^{-1}(C)} \right)^2 \right)
\: .
\end{align*}
We intend to prove an exponential decrease with $x$. In order to have it, we will set $t_k$ such that the exponential is equal to 1 for $x=0$, and then decreases as $x$ grows.
Let then $t_k$ be such that $C \le F_k(t_k, t_k \Delta_k^2)$. It exists as soon as  $C \ge F_k(0,0)$ (where the later is non-positive for specific algorithms we will consider). For all $t \ge t_k$, $F_k(t, t \Delta_k^2) \ge F_k(t_k, t_k \Delta_k^2) \ge C$, which leads to $\sqrt{\Delta_k^2 t} - \sqrt{F_k(t, \cdot)^{-1}(C)} \ge 0$. Note that since $F_k$ is non-decreasing in the first variable we have $F_k(t+x, \cdot)^{-1}(C) \le F_k(t, \cdot)^{-1}(C)$ for all $t,x \ge 0$, and 
\begin{align*}
\delta_{t_k+x, k, C}
&= \exp\left( - \left(\sqrt{\Delta_k^2(t_k+x)} - \sqrt{F_k(t_k+x, \cdot)^{-1}(C)} \right)^2 \right)
\\
&\le \exp\left( - \left(\sqrt{\Delta_k^2(t_k+x)} - \sqrt{F_k(t_k, \cdot)^{-1}(C)} \right)^2 \right)
\\
&\le \exp\left( - \Delta_k^2 \left(\sqrt{t_k+x} - \sqrt{t_k} \right)^2 \right)
\: .
\end{align*}
Let $Y_k = \max(\sqrt{\tau_k(C)}, \sqrt{t_k})$; we have proved that for all $x \ge 0$,
\begin{align*}
\mathbb{P}(Y_k > \sqrt{t_k + x})
&\le \exp\left( - \Delta_k^2 \left(\sqrt{t_k+x} - \sqrt{t_k} \right)^2 \right)
\: .
\end{align*}
By setting $x = 2 \sqrt{\lambda t_k} + \lambda$ for $\lambda \ge 0$, we get
$\mathbb{P}(Y_k > \sqrt{t_k} + \lambda) \le \exp\left( - \Delta_k^2 \lambda^2  \right)$ .
\end{proof}

\begin{lemma}\label{lem:sum_tau_tail} For all $C>0$, $t_{k}$ such that $F(t_{k}, t_{k}\Delta_k^2) \ge C$.
	\begin{equation*}
	\mathbb{P}(\sum_k \tau_k(C) \ge T)
	\le e \times \exp\left(-\frac{\frac{T}{2} - \sum_k t_{k}}{\sum_k 1/\Delta_k^2}\right).
	\end{equation*}
\end{lemma}
\begin{proof}
	Rewrite the event $\{\sum_k\tau_k(C) > T\}$ using $Y_k = \max(\sqrt{\tau_k(C)}, \sqrt{t_k})$, so that:
	\begin{align*}
	\mathbb{P}\left(\sum_k\tau_k(C) > T\right)
	&= \mathbb{P}\left(\sum_k ((Y_k - \sqrt{t_k}) + \sqrt{t_k})^2 > T\right)
	\\
	&\le \mathbb{P}\left(2\sum_k (Y_k - \sqrt{t_k})^2 +2 \sum_k t_k > T\right)
	\\
	&\le \mathbb{P}\left(\sum_k (Y_k - \sqrt{t_k})^2  > T/2 - \sum_k t_k\right) \label{eq:deviation_T/2}\numberthis
	\: .
	\end{align*}
	
	We now apply Lemma~\ref{lem:sum_exponentials} to $(Y_k - \sqrt{t_k})^2$, which verifies $\mathbb{P}((Y_k - \sqrt{t_k})^2 \ge x) \le \exp(- \Delta_k^2 x)$. From Equation \eqref{eq:deviation_T/2}, we obtain
	\begin{align*}
	\mathbb{P}(\sum_k\tau_k(C) > T)
	\le \mathbb{P}(\sum_k (Y_k - \sqrt{t_k})^2  > T/2 - \sum_k t_k)
	&\le e \times \exp \left( - \frac{T/2 - \sum_k t_k}{\sum_k 1 / \Delta_k^2}\right)
	\: .
	\end{align*}
\end{proof}

\paragraph{Remark} We can actually derive a tighter bound than \eqref{eq:deviation_T/2}
\begin{align*}
\mathbb{P}(\sum_k\tau_k(C) > T)
&= \mathbb{P}(\sum_k ((Y_k - \sqrt{t_k}) + \sqrt{t_k})^2 > T)
\\
&\le \mathbb{P}\left( \left(\sqrt{\sum_k (Y_k - \sqrt{t_k})^2} + \sqrt{\sum_k t_k}\right)^2 > T\right)
\\
&= \mathbb{P}\left(\sum_k (Y_k - \sqrt{t_k})^2  > (\sqrt{T} - \sqrt{\sum_k t_k})^2\right) \label{eq:deviation_sqrt(T)}
\: .
\end{align*}
Roughly speaking, to get it, just write $\| (Y - \sqrt{t}) + \sqrt{t} \|^2 \le (\| Y - \sqrt{t} \| + \|\sqrt{t} \|)^2$. To get Equation \eqref{eq:deviation_T/2}, we further use $(\| Y - \sqrt{t} \| + \|\sqrt{t} \|)^2 \le 2\| Y - \sqrt{t} \|^2 + 2\|\sqrt{t} \|^2$.
In the case of APT (at least) it leads to the \emph{same} final bound on the algorithm because when we optimize further down, we set $\sum_k t_k = T/4$ no matter which of these inequalities we use, value for which resulting exponents are equal.

\begin{corollary}
	Suppose now that there is a set $S_C$ such that for $k \notin S_C$, $\mathbb{P}(\tau_k(C) > t_k) = 0$. Then we can refine Lemma~\ref{lem:sum_tau_tail} to
	\begin{align*}
	\mathbb{P}(\sum_k \tau_k(C) > T)
	\le \mathbb{P}(\sum_{k \in S_C} \tau_k(C) > T - \sum_{k\notin S_C} t_k)
	&\le e \times \exp\left(-
	\frac{(T - \sum_{k\notin S_C} t_k)/2 - \sum_{k \in S_C} t_k}{\sum_{k \in S_C} 1 /\Delta_k^2}\right)
	\: .
	\end{align*}
\end{corollary}

\subsection{When an arm index is large, the probability of mistake is small} 
\label{sub:when_arm_well_explored}

The goal of this section is to bound $\mathbb{P}(\hat{s}_k \ne s_k \land \exists T' \in [T], I_{N_{T'}}^k \ge C)$. We define the random variable $E_{k,t} = \mathbb{I}\{(\hat{\mu}_{k,t} - \theta)(\mu_k - \theta)<0\}$; it is equal to 1 iff there is an error on the sign at time $t$. The algorithm makes a mistake on arm $k$ is $E_{k,t_{\max}} = 1$ since it returns the sign at that time.

\begin{theorem}\label{lem:proba_error_tail}
	The algorithm using $F_k$ for its index definition verifies
	\begin{align*}
	\mathbb{P}(\hat{s}_k \ne s_k \land \exists T' \in [T], I_{N_{T'}}^k \ge C)
	&\le \mathbb{P}(\exists T' \in [T], E_{k,T'} = 1 \land I_{N_{T'}}^k \ge C)
	\\
	&\le T \cdot \inf \{e^{-n_k \Delta_k^2} \mid F_k(n_k, n_k \Delta_k^2) < C\} \: .
	\end{align*}
\end{theorem}
\begin{proof}
	We use Lemma~\ref{lem:n_C_inequality} (below): find $n_k$ as large as possible such that $F_k(n_k, n_k \Delta_k^2) < C$. Then, since the algorithm returns the sign of the arm at the time at which its index was maximal, we get $\mathbb{P}(\hat{s}_k \ne s_k \land \exists T' \in [T], I_{N_{T'}}^k \ge C) \le \mathbb{P}(\exists T' \in [T], E_{k,T'} = 1 \land I_{N_{T'}}^k \ge C) \le T\cdot e^{-n_k \Delta_k^2}$.
\end{proof}

\begin{lemma}\label{ineq:uniform_gap_bound}  For any $\delta_k \in (0,1)$, with probability at least $1 - \delta_k$, and for all $n \in [T]$, it holds
	\begin{equation*}
	\sqrt{N_{k,t}}(\hat{\mu}_t^k - \mu^k) \le \sqrt{\log \left(\frac{T}{\delta_k}\right)}.
	\end{equation*}
\end{lemma}
\begin{proof}
	This is a direct implication of Hoeffding's inequality with a union bound for time-uniformity.
\end{proof}

Define $n_k = \frac{1}{\Delta_k^2}\log \left(\frac{T}{\delta_k}\right)$.
Consider the following  three facts (their definition will be useful for the following proofs):
\begin{enumerate}[nosep]
	\item If the concentration holds, then $\sqrt{N_{k,t}}(\hat{\mu}_t^k - \mu^k) \le \sqrt{n_k}\Delta_k$. 
	\item If there is a mistake at time $t$, then we have
	\begin{enumerate}
		\item $\hat{\mu}_t^k - \mu^k \ge \Delta_k$.
		\item $(\hat{\mu}_t^k - \mu^k)^2 \ge \hat{\Delta}_{N_{k,t},k}^2 + \Delta_k^2$.
	\end{enumerate}
	\item If $I_{N_{k,t}}^k > C$ then $F_k(N_{k,t}, N_{k,t} \hat{\Delta}_{t,k}^2) > C$ and $N_{k,t}\hat{\Delta}_{t,k}^2 \ge F_k(N_{k,t}, \cdot)^{-1}(C)$.
\end{enumerate}

\begin{lemma}\label{lem:concentration_and_mistake}
	If at time $t$, concentration holds and there is a mistake (1 and 2 are true), then
	$N_{k,t} \le n_k$ and $F_k(N_{k,t}, (n_k - N_{k,t}) \Delta_k^2) \ge I_{N_{k,t}}^k$.
\end{lemma}
\begin{proof}
	First point: combine 1 and 2(a).
	Second point: use 1, then 2(b), then the definition of $I_{N_{k,t}}^k$:
	\begin{align*}
	n_k \Delta_k^2 \ge N_{k,t} (\hat{\mu}_t^k - \mu^k)^2 \ge N_{k,t} \hat{\Delta}_{t,k}^2 + N_{k,t} \Delta_k^2 \ge F_k(N_{k,t}, \cdot)^{-1}(I_{N_{k,t}}^k) + N_{k,t} \Delta_k^2 \: .
	\end{align*}
\end{proof}

\begin{lemma}\label{lem:n_C_inequality}
	If at time $t$, all three ``if'' are true, then $F_k(n_k, n_k \Delta_k^2) \ge C$.
\end{lemma}
\begin{proof}
	Use the monotonicity of $F_k$ in the inequality of Lemma~\ref{lem:concentration_and_mistake}.We have $N_{k,t} \le n_k$ and $n_k - N_{k,t} \le n_k$.
\end{proof}


\subsection{Examples}\label{app:Examples}
In this section, we explicit Theorem~\ref{thm:expected_errors_bound} for certain algorithms from the literature and for our algorithm, FWT.
\subsubsection{APT (\cite{locatelli2016optimal})} 
This algorithm (in its variant that stops at $t_{\max}$) corresponds to $F(n,x) = x$. To apply Theorem~\ref{thm:expected_errors_bound} we find that $t_j(C_k) = \frac{C_k}{\Delta_j^2}$ is solution, then:
\begin{align*}
\mathbb{E}[L_T]
&\le \sum_{k =1}^K e a_k \exp\left(- \left(\frac{T/2}{\sum_{j =1}^K 1/\Delta_j^2}- C_k \right)\right) + T\cdot a_k \exp \left(- C_k \right)
\end{align*}
An optimal $C_k$ is such that
\begin{align*}
e \cdot \exp\left(-
\frac{\frac{T}{2}}{\sum_{j =1}^K 1/\Delta_j^2}+ C_k \right)	=T\cdot \exp( -C_k)
\end{align*} 
Then the bound becomes:
\begin{align*}
\mathbb{E}[L_T] \le 2\sqrt{e\cdot T}\sum_{k =1}^K a_k  \exp\left(-\frac{1}{4}\frac{T}{\sum_{k=1}^K 1/\Delta_k^2}\right)
\end{align*}

\subsubsection{LSA (\cite{tao2019thresholding})}
This algorithm corresponds to $F(n,x) = x + \log n$, the stopping time $t_j(C_k)$ is solution to $t\Delta_j^2 + \log t = C_k $. This equation has a closed form solution: $t_j(C_k)=\frac{1}{\Delta_j^2}W(\Delta_j^2 \exp(C_k))$, the loss bound becomes
%
\begin{align*}
	\mathbb{E}[L_T] \le& \sum_{k =1}^K e\cdot a_k \exp\left(-
	\left(\frac{\frac{T}{2} - \sum_{j=1}^K \frac{1}{\Delta_j^2}W(\Delta_j^2 \exp(C_k))}{\sum_{j =1}^K 1/\Delta_j^2} \right)\right) +T a_k \exp \left(- W(\Delta_k^2 \exp(C_k))\right)
\end{align*}
We can use an inequality on the lambert function (\cf~Corollary 2.4 in \cite{hoorfar2008inequalities}), For all $x \geq e$ we have
\begin{equation*}
		\log x-\log \log x \leq W(x) \leq \log x-\log \log x+\log \left(1+e^{-1}\right),
\end{equation*}
this entails that if $\forall k,j \quad C_k + \log \Delta_j^2 \ge 1$ we obtain the more accurate bound
\begin{align}
\mathbb{E}[L_T] \le& \sum_{k =1}^K (1+e)\cdot a_k \exp\left(-
\left(\frac{\frac{T}{2} + \sum_{j=1}^K \frac{1}{\Delta_j^2}(\log \frac{1}{\Delta_j^2} + \log(C_k + \log\Delta_j^2))}{\sum_{j =1}^K 1/\Delta_j^2} \right) + C_k\right) \nonumber\\
 &+\sum_{k =1}^K\frac{T (C_k+\log\Delta_k^2)}{\Delta_k^2} a_k \exp \left(- C_k\right).
 \label{eq:upper_bound_LSA}
\end{align}

\subsubsection{FWT (our algorithm)}

Our algorithm corresponds to $F_k(n,x) = \max(1,x) - \log \max(1,x) + \log n - \log a_k$. In order to find the times $t_j(C_k)$ of Theorem~\ref{thm:expected_errors_bound} we solve the equation:
\begin{align*}
\max(1,n \Delta_k^2) - \log\max(1,n \Delta_k^2) + \log (n \Delta_k^2) = C + \log a_k + \log \Delta_k^2\: .
\end{align*}
Let $D_k = C + \log a_k + \log \Delta_k^2$. We want a solution to
$
\max(1,x) - \log \max(1,x) + \log x = D_k
$.
The function on the left, which we now denote by $\mathcal I$, is increasing and bijective from $\mathbb{R}^+$ to $\mathbb{R}$. 
\begin{itemize}
	\item If $D_k \ge 1$, $\mathcal I (D_k) = D_k$.
	\item If $D_k \le 1$, $\mathcal I (e^{D_k - 1}) = D_k$.
	\item For all $D_k > 0$, $\mathcal I(e^{D_k - 1}) \ge D_k$.
\end{itemize}
Moreover, we have $F_k(a_k e^{C-1})\ge C$, it comes $\mathbb{P}(\tau_k(C)>a_k e^{C-1})=0$.\\
Let $C_k>0$ and $S_k \subseteq \{j \in [K] \ | \ C_k + \log a_j\Delta_j^2 \ge 1 \}$. Let $S'$ be a set such that for all $k \in S'$, $k \in S_k$.
\begin{align*}
\mathbb{E}[L_T]
&\le \sum_{k\notin S'} a_k
+ \sum_{k\in S'} a_k\mathbb{P}(\overline{\mathcal F_{C_k}}) + \sum_{k\in S'} a_k\mathbb{P}(\overline{\mathcal E_k(T)} \cap \mathcal F_{C_k})
\\
&\le \sum_{k\notin S'} a_k
+ e \sum_{k\in S'} a_k \exp\left(- \frac{\frac{1}{2}(T - \sum_{j \notin S_k} a_j e^{C_k-1}) - \sum_{j \in S_k} \frac{1}{\Delta_j^2}\log a_j\Delta_j^2}{\sum_{j \in S_k} 1/\Delta_j^2} + C_k\right)
\\ &\qquad + T \sum_{k\in S'} a_k \exp \left( - C_k - \log a_k\Delta_k^2 \right)
\: .
\end{align*}

\paragraph{Large $T$} The value of $C_k$ which equalizes the two terms indexed by $k$ verifies
\begin{align*}
C_k = \frac{1}{2} \frac{\frac{1}{2}(T - \sum_{j \notin S_k} a_j e^{C_k-1}) - \sum_{j \in S_k} \frac{1}{\Delta_j^2}\log a_j\Delta_j^2}{\sum_{j \in S_k} 1/\Delta_j^2} - \frac{1}{2}\log a_k \Delta_k^2 + \frac{1}{2}\log \frac{T}{e}
\: .
\end{align*}

The latter can be chosen if $T$ is big enough such that $S_k = [K]$ for all arms, this is the case if $T \ge 2 \sum_{j=1}^K \frac{1}{\Delta_j^2} (2 + \log \frac{a_j\Delta_j^2 \max_i a_i \Delta_i^2}{(\min_k a_k \Delta_k^2)^2}- \log \frac{T}{e^3}) $, we get the bound
\begin{align*}
\mathbb{E} [L_T]
\le 2 \sqrt{e T} \sum_k a_k \exp \left( - \frac{1}{2} \frac{T/2 - \sum_{j} \frac{1}{\Delta_j^2}\log \frac{a_j\Delta_j^2}{a_k \Delta_k^2}}{\sum_{j} 1/\Delta_j^2}\right)
\end{align*}
Up to a $1/4$ factor, this is the exponent of the optimal non-adaptive oracle (\cf~Eq.~\ref{ineq:Objective}).

\paragraph{General $T$}
We choose a set $S \subseteq [K]$ and set $C_k = C$, a common value still to be determined, for all $k \in S'$. Then for all $k \in S'$, we set $S_k = S$. We impose $C \ge 1 + \max_{j \in S}\log \frac{1}{a_j \Delta_j^2}$, such that the condition $S_k \subseteq \{j \in [K] \ | \ C_k + \log a_j\Delta_j^2 \ge 1 \}$ is verified. We get that for all $S' \subseteq S \subseteq [K]$ and $C \ge 1 + \max_{j \in S}\log \frac{1}{a_j \Delta_j^2}$~,
\begin{align*}
\mathbb{E}[L_T]
&\le \sum_{k\notin S'} a_k
+ e \sum_{k\in S'} a_k \exp\left(- \frac{\frac{1}{2}(T - \sum_{j \notin S} a_j e^{C-1}) + \sum_{j \in S} \frac{1}{\Delta_j^2}\log \frac{1}{a_j\Delta_j^2}}{\sum_{j \in S} 1/\Delta_j^2} + C \right)
\\ &\qquad + T \sum_{k\in S'} a_k \exp \left( - C - \log a_k \Delta_k^2 \right)
\: .
\end{align*}

\section{Extention to sum-of-gaps}
\label{ap:sum_of_gaps}

The global loss we investigate in this section is \begin{equation*}
	L_T = \sum_k \Delta_k E_k.
\end{equation*}

First we write the Frank-Wolfe index: $\argmin_k N_{k,t} \hat{\Delta}_{k,t}^2 -\frac{3}{2}\log\left(N_{k,t} \hat{\Delta}_{k,t}^2\right)+\frac{3}{2}\log\left(N_{k,t}\right)$, then we slightly modify it to comply with Assumption~\ref{ass:index_shape} (see explanation above Eq.~\ref{def:FWT}): \begin{equation*}
	\ I_{N_{k,t}}^k = \max(\frac{3}{2},N_{k,t} \hat{\Delta}_{k,t}^2) - \frac{3}{2}\log \max(\frac{3}{2},N_{k,t} \hat{\Delta}_{k,t}^2) + \frac{3}{2}\log N_{k,t}.
\end{equation*}
This corresponds to the function $F(n,x) = \frac{3}{2}\log n + \max(\frac{3}{2},x) - \frac{3}{2}\log \max(\frac{3}{2},x)$, used for all arms. 
\\
Solving $F(n, n \Delta_k^2) = C$ gives rise to two cases:
\begin{itemize}[nosep]
	\item $n = \frac{1}{\Delta_k^2}(C + \frac{3}{2}\log\Delta_k^2)$ if $C + \frac{3}{2}\log\Delta_k^2 \ge \frac{3}{2}$,
	\item $n = \frac{3}{2}\exp(\frac{2}{3}C - 1)$ otherwise. In that case, $n \le \frac{3}{2}\frac{1}{\Delta_k^2}$.
\end{itemize}

Also, $n = \frac{3}{2}\exp(\frac{2}{3}C - 1)$ is solution to $F(n, 0) = C$.

Consider $C_k>0$, let $S_k = \{j \in [K] \ | \ C_k + \frac{3}{2}\log \Delta_j^2 \ge \frac{3}{2} \}$ and $S'$ be a set such that for all $k \in S'$, $k \in S_k$, then
\begin{align*}
\mathbb{E}[L_T]
&\le \sum_{k\notin S'} \Delta_k
+ \sum_{k\in S'} \Delta_k\mathbb{P}(\overline{\mathcal F_{C_k}}) + \sum_{k\in S'} \Delta_k\mathbb{P}(\overline{\mathcal E_k(T)} \cap \mathcal F_{C_k})
\\
&\le \sum_{k\notin S'} \Delta_k
+ e \sum_{k\in S'} \Delta_k \exp\left(- \frac{\frac{1}{2}(T - \sum_{j \notin S_k} \frac{3}{2}\Delta_j e^{\frac{2}{3}C_k-1}) + \sum_{j \in S_k} \frac{3}{2}\frac{1}{\Delta_j^2}\log \frac{1}{\Delta_j^2}}{\sum_{j \in S_k} 1/\Delta_j^2} + C_k\right)
\\ &\qquad + \sum_{k\in S'} T \Delta_k \exp \left( - C_k - \frac{3}{2}\log \Delta_k^2 \right)
\end{align*}

\paragraph{Large $T$} The values of $C_k$ that optimize the r.h.s of the previous inequality verify:
\begin{equation*}
	C_k = \frac{1}{2}\frac{\frac{1}{2}(T - \sum_{j \notin S_k} \frac{3}{2}\Delta_j e^{\frac{2}{3}C_k-1}) + \sum_{j \in S_k} \frac{3}{2}\frac{1}{\Delta_j^2}\log \frac{1}{\Delta_j^2}}{\sum_{j \in S_k} 1/\Delta_j^2}-\frac{1}{2}\log \Delta_k^3 + \frac{1}{2}\log\frac{T}{e}
\end{equation*}
If $T$ is big enough such that $S_k = [K]$ for all arms, which happens once $T\ge 2 \sum_{j=1}^k \frac{1}{\Delta_j^2}\left(3 + 3 \log\frac{\Delta_j \max_{i}\Delta_i}{(\min_{k}\Delta_i)^2}-\log\frac{T}{e}\right)$, we get the bound:
\begin{align*}
\mathbb{E}[L_T]
&\le 2 \sqrt{e T} \sum_{k} \Delta_k \exp\left(- \frac{1}{2}\frac{\frac{T}{2} + \sum_{j} \frac{3}{2}\frac{1}{\Delta_j^2}\log \frac{\Delta_k^2}{\Delta_j^2}}{\sum_{j} 1/\Delta_j^2}\right)
\: .
\end{align*}

\section{Upper-bounds comparison}

\begin{figure}[!h]
\begin{center}
  \includegraphics[width=.45\textwidth]{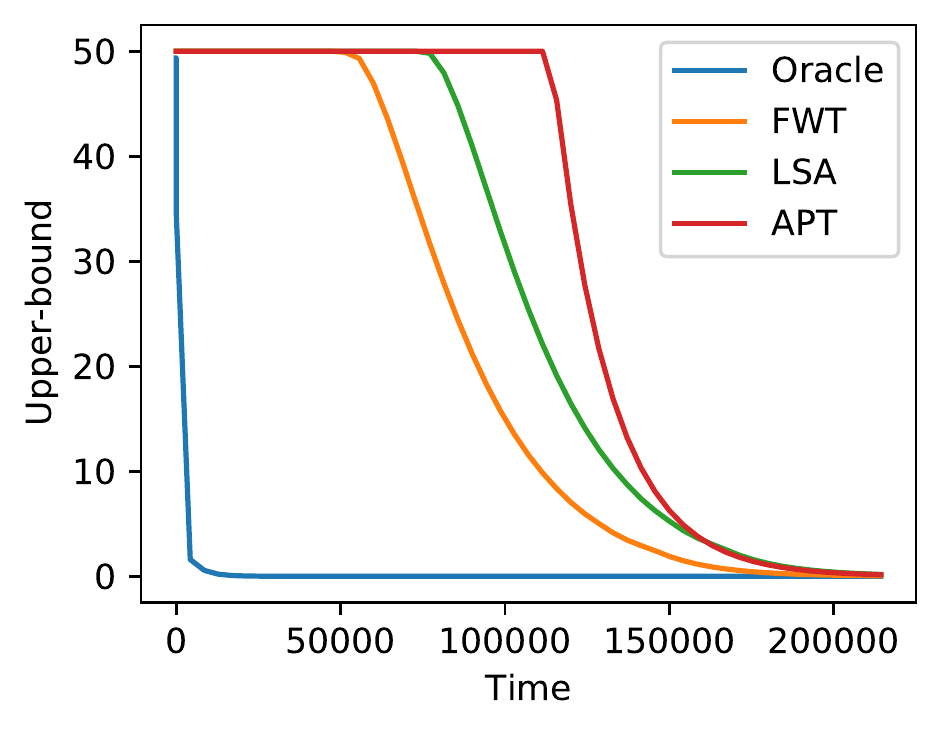} \quad
  \includegraphics[width=.45\textwidth]{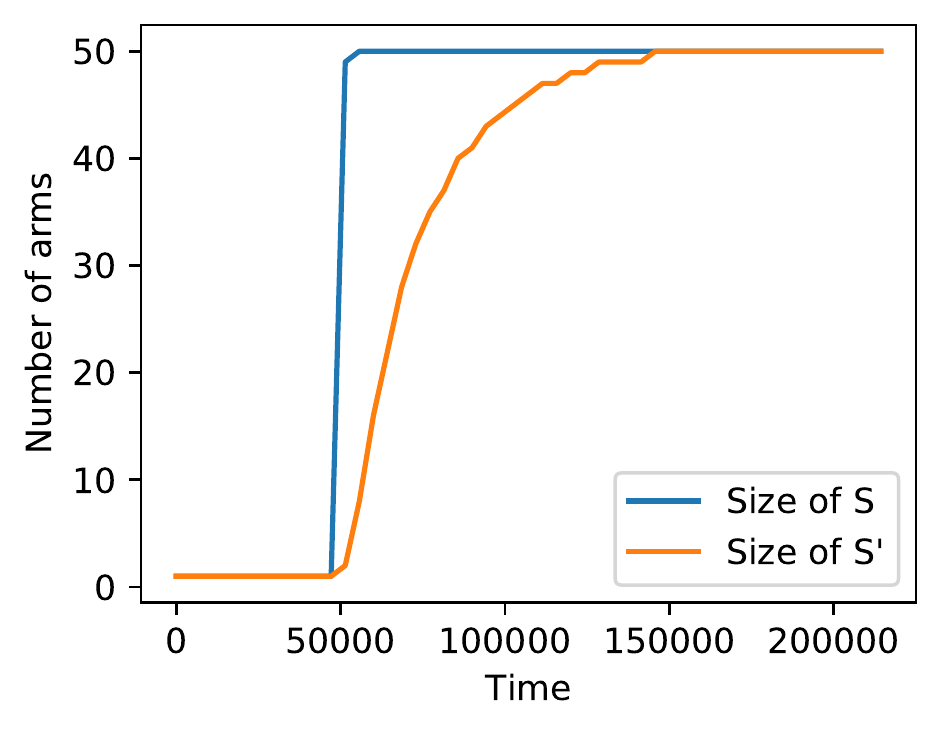} 
\end{center}
\caption{[left] Comparison of the upper-bounds of Corollaries~\ref{cor:APT_upperbound},~\ref{cor:FWTupperbound} and that of the optimal non-adaptive oracle of Eq.~\eqref{eq:OptimalRegret} (blue) when the gaps are of the form $\Delta_i = i/K$. [right] Evolution over time of the size of the optimal sets $S$ (blue) and $S'$ (orange) that minimize the bound of Corollary~\ref{cor:FWTupperbound}. }
\label{fig:upperbounds_linear}
\end{figure}

Figure~\ref{fig:upperbounds_linear} compares the upper-bounds of Corollary~\ref{cor:APT_upperbound} (APT), Equation~\eqref{eq:upper_bound_LSA} (LSA), and Corollary~\ref{cor:FWTupperbound} (FWT) for the particular case $\Delta_i = i/K$ and $a_i = 1$, for all $i = 1,\dots, K$ and $K=50$. We observe a behavior similar to that of Figure~\ref{fig:upperbounds}.

\section{Additional Experiments}
\label{app:experiments}

In this section, we illustrate on synthetic data the performance of APT, LSA, and FWT. The implemented algorithms respectively correspond to Algorithm~\ref{alg:index-based} with the following choices:
\begin{align}
	F(n,x;1) & = x   \tag{APT} \\
	F(n,x;1) & = x + \log(n) \tag{LSA} \\
	F(n,x;1) & = (1+ \sqrt{x})^2 - \log (1+ \sqrt{x})^2 + \log(n) \tag{FWT}
\end{align}
Note that we used a slighly different version for APT than the one proposed in the analysis $F(n,x;1) = \max\{1,x\} - \log \max\{1,x\} + \log(n)$. The analysis and experiments work similarly for both versions. But the $(1+ \sqrt{x})^2$ version performs slightly better empirically while the $\max\{1,x\}$ version provides cleaner theoretical results. The experiments are averaged over 500 runs and consider arm distributions of the form $\nu_k = \mathcal{N}(\mu_k, 1)$. The gaps are thus $\Delta_k = |\mu_k|$, for $k = 1,\dots,K$. The performance criterion is the sum of errors defined in Equation~\ref{ineq:Objective} with $a_k = 1$. The experiments were run on a personal laptop with Intel Core i5, Dual Core, $3.1$ GHz.

Since most of the tested experiments obtained similar performance, we only provide the results of a few experiments. Although our theoretical upper bounds are slighly better for FWT, LSA and FWT generally have similar performance, while APT underperforms. This last point is not surprising since, although we provide in Corollary~\ref{cor:APT_upperbound} an upper bound for APT that appears asymptotically similar to those of LSA and FWT, APT was not designed to minimize the sum of errors. APT was made to minimize the probability of making at least one error and thus focuses too much on arms with very small gaps that are very difficult to classify.

\begin{figure}
\begin{center}
	\includegraphics[width=.45\textwidth]{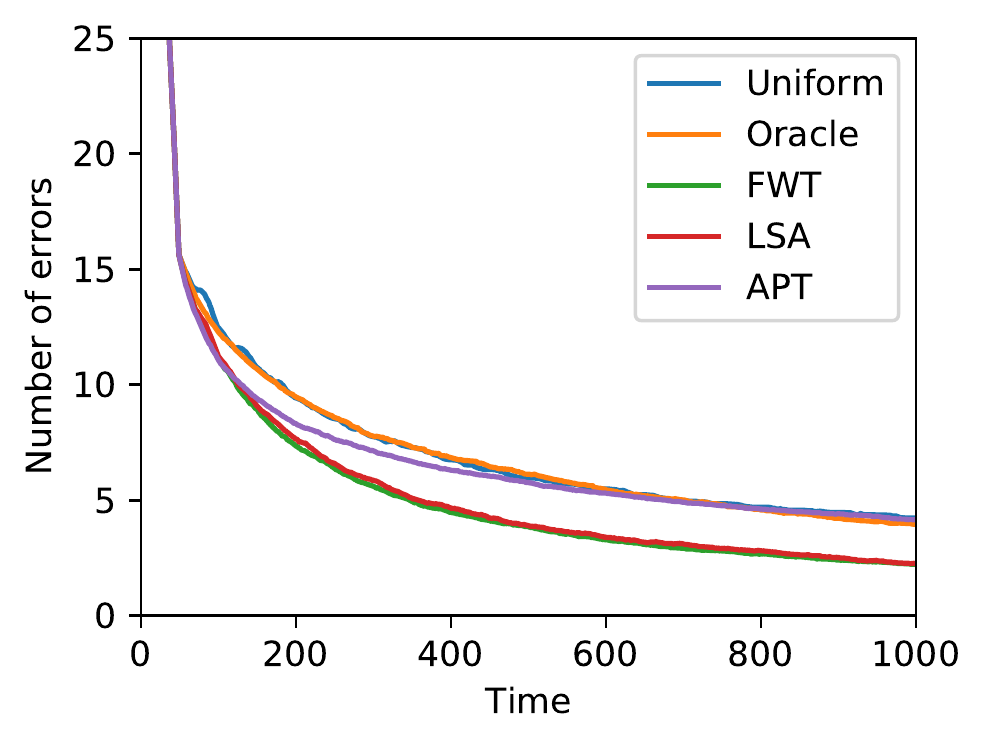}\quad
	\includegraphics[width=.45\textwidth]{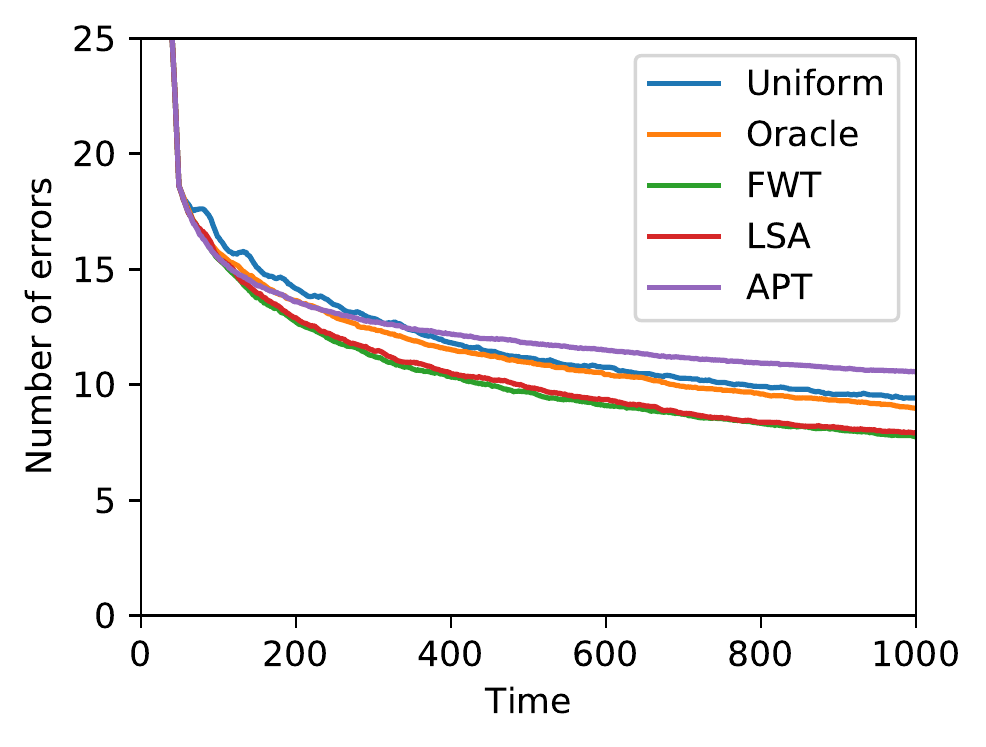}
\end{center}
	\caption{Sum of errors over time of the different algorithms when $\mu_k = \frac{(-1)^k k}{K}$ [left] and  $\mu_k = \frac{(-1)^k k^2}{K^2}$ [right] for $K=50$ arms.}
	\label{fig:performance}
\end{figure}

\begin{figure}
\begin{center}
	\includegraphics[width=.45\textwidth]{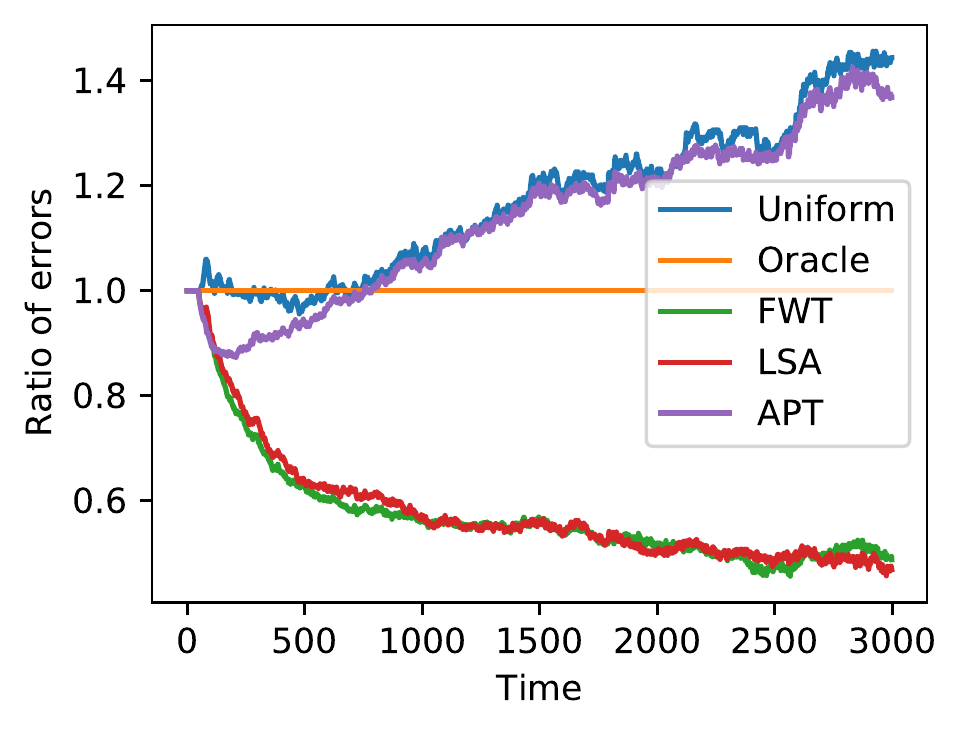}\quad
	\includegraphics[width=.45\textwidth]{img/improvement_over_oracle_square.pdf}
\end{center}
	\caption{Ratio of improvement with respect to the non adaptive oracle sampling when $\mu_k = \frac{(-1)^k k}{K}$ [left] and  $\mu_k = \frac{(-1)^k k^2}{K^2}$ [right] for $K=50$ arms.}
	\label{fig:improv_unif}
\end{figure}

Figure~\ref{fig:performance} shows the performance of the algorithms together with the non-adaptive oracle of Section~\ref{sec:non_adaptive_algorithms}. Figure~\ref{fig:improv_unif} plots the ratio of error with respect to the non-adaptive oracle. Interestingly, in all of our experiments, APT and FWT perform better than it.

\begin{figure}
\begin{center}
	\includegraphics[width=\textwidth]{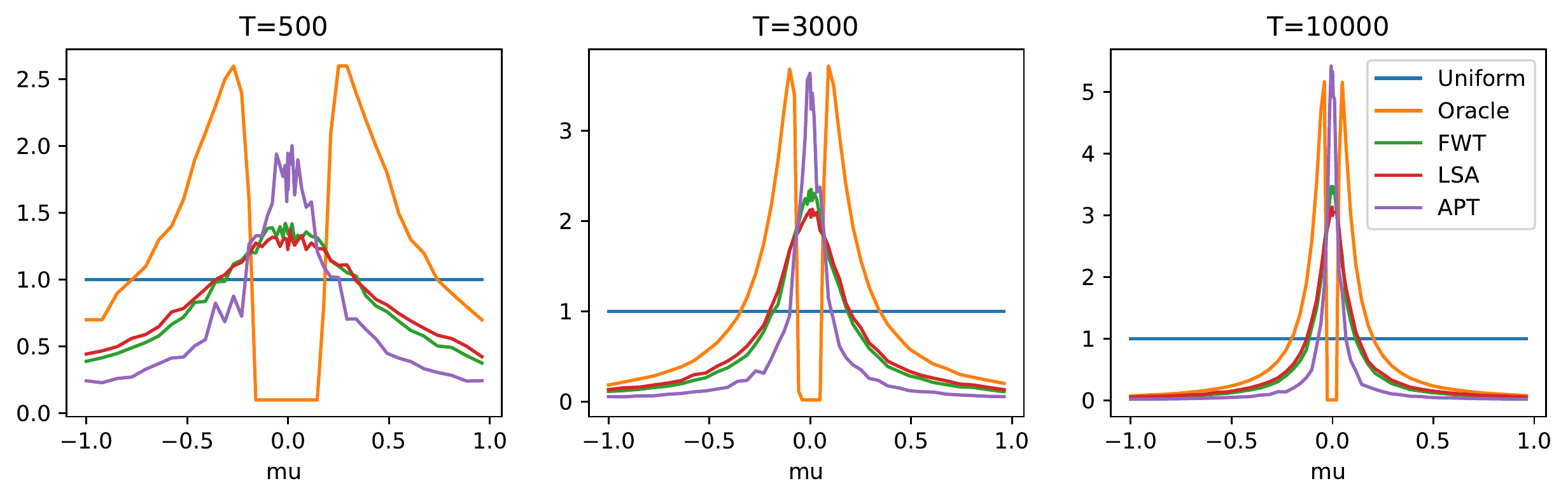}
	\caption{Optimal sampling distribution and empirical sampling distribution with respect to $\mu$ when $\mu_k = \frac{(-1)^k k^2}{K^2}$ for $K=50$ arms.}
	\label{fig:distribution_linear}
\end{center}
\end{figure}

\begin{figure}
\begin{center}
	\includegraphics[width=\textwidth]{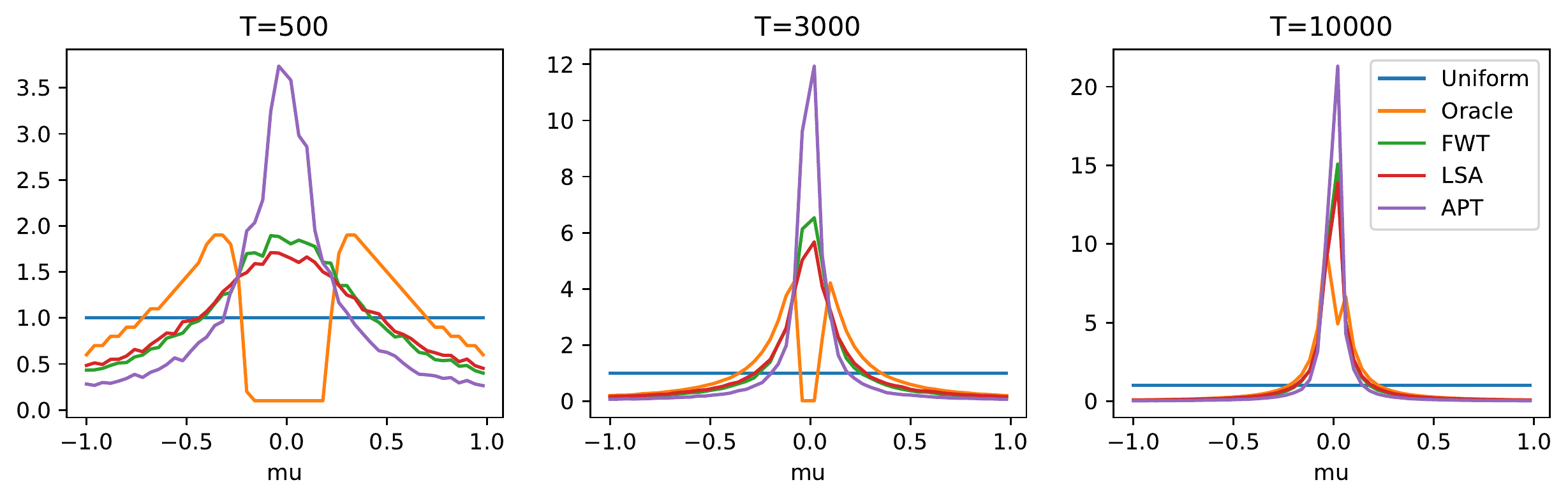}
	\caption{Optimal sampling distribution and empirical sampling distribution with respect to $\mu$ when $\mu_k = \frac{(-1)^k k}{K}$ for $K=50$ arms.}
	\label{fig:distribution_square}
\end{center}
\end{figure}

Figures~\ref{fig:distribution_linear} and \ref{fig:distribution_square} represent the optimal non-adaptive sampling distribution if the means were known and the empirical sampling distribution of the algorithms for different numbers of iterations. As we can see, for the initial phase, the arms that are closest to the threshold should ideally not be drawn. Yet, as our lower bound in Section~\ref{sub:lower_bound_on_the_number_of_pulls_of_arms_close_to_zero} illustrates, this is not possible for sequential algorithms. All arms must be sampled. We can see that this is indeed the case for all algorithms: the closer the arms are to the threshold, the more likely they are to be sampled. 

\end{document}